\documentclass[preprint,12pt]{elsarticle}
\usepackage{amsfonts}
\usepackage{amsmath}
\usepackage{psfrag}
\usepackage{graphicx}
\usepackage{amssymb}
\usepackage{subfigure}
\usepackage{algorithmic}
\usepackage{algorithm}
\usepackage{color}
\usepackage{cases}
\usepackage{tikz}

\usepackage{natbib}
\usepackage{float}
\usepackage{circuitikz}

\newtheorem{thm}{Theorem}

\newproof{pot}{Proof}

\newtheorem{defi}{Definition}

\newtheorem{corollary}{Corollary}
\newtheorem{remark}{Remark}
\newtheorem{example}{Example}

\allowdisplaybreaks[4]

\begin{document}
	
\begin{frontmatter}
\title{Zeroing neural dynamics solving time-variant complex conjugate matrix equation $X(\tau)F(\tau)-A(\tau)\overline{X}(\tau)=C(\tau)$}

\author[1,2]{Jiakuang He}
\author[2]{Dongqing Wu\corref{cor1}}
\ead{rickwu@zhku.edu.cn}

\address[1]{School of Information Science and Technology, Zhongkai University of Agriculture and Engineering, Guangzhou 510225, P.
	R. China}
\address[2]{School of Mathematics and Data Science, Zhongkai University of Agriculture and Engineering, Guangzhou 510225, P.
	R. China}
\cortext[cor1]{Corresponding author}
	
\begin{abstract}
Complex conjugate matrix equations (CCME) are important in computation and antilinear systems. Existing research mainly focuses on the time-invariant version, while studies on the time-variant version and its solution using artificial neural networks are still lacking. This paper introduces zeroing neural dynamics (ZND) to solve the earliest time-variant CCME. Firstly, the vectorization and Kronecker product in the complex field are defined uniformly. Secondly, Con-CZND1 and Con-CZND2 models are proposed, and their convergence and effectiveness are theoretically proved. Thirdly, numerical experiments confirm their effectiveness and highlight their differences. The results show the advantages of ZND in the complex field compared with that in the real field, and further refine the related theory.
\end{abstract}

\end{frontmatter}



\section{Introduction}\label{sec.intro}
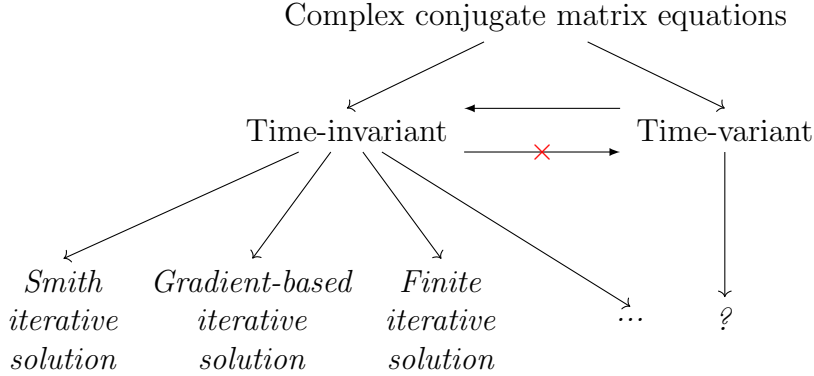
\begin{figure}[htbp]
	\begin{tikzpicture}[
		level 1/.style={sibling distance=5cm},
		level 2/.style={sibling distance=2.5cm},
		edge from parent/.style={  
			->,  
			draw,  
			edge from parent path={(\tikzparentnode) -- (\tikzchildnode.north)}  
		},  
		every node/.style={shape=rectangle,  
			rounded corners,   
			align=center  
			},
		myarrow/.style={  
				->,  
				>=latex,  
				shorten >=2pt,  
				shorten <=2pt,  
				draw  
			}
		]
		\node {Complex conjugate matrix equations}
		child {node(time-invariant) {Time-invariant}
			child {node[yshift=-10mm] {\textit{Smith}\\\textit{iterative}\\\textit{solution}}}
			child {node[yshift=-10mm] {\textit{\textit{Gradient-based}}\\\textit{iterative}\\\textit{solution}}}
			child {node[yshift=-10mm] {\textit{Finite}\\\textit{iterative}\\\textit{solution}}}
			child {node[yshift=-10mm] {\textit{...}}}
		}
		child {node(time-variant)  {Time-variant}
			child {node[yshift=-10mm] {\textit{?}}}
		};
		\draw[myarrow] (time-variant.north west) -- (time-invariant.east |- time-variant.north west);
		\draw[myarrow] (time-invariant.south east)-- (time-variant.west |- time-invariant.south east)node[pos=0.5,anchor=center,color=red] {$\times$};  
	\end{tikzpicture}
	\caption{Solving methods of CCME.}
	\label{fig.compare.solve}
\end{figure}

Complex conjugate matrix equations (CCME) 
\cite{bevisConsimilarityMatrixEquation1987,bevisMatrixEquationTextbackslashtextbackslashA1988,jiangSolutionsMatrixEquations2003,
Wu2017}
are matrix equations containing unknown matrices and their complex conjugate matrices. Among them, complex conjugate matrices are linked to Lyapunov equations in the complex field 
\cite{lvWsbpFunctionActivated2018,uhligZhangNeuralNetworks2024},
Hermitian matrices
\cite{golubMatrixComputations4th2013},
etc. Because the related matrix equations containing complex conjugate matrices have certain refinements and supplements to the control theory like antilinear systems
\cite{bevisConsimilarityMatrixEquation1987,bevisMatrixEquationTextbackslashtextbackslashA1988,jiangSolutionsMatrixEquations2003,Wu2017}
and numerical computations for matrices
\cite{Wu2017,wuSolutionsMatrixEquations2006,wuMatrixEquationsAXF2009,songEfficientAlgorithmSolving2011,wuConjugateProductComplex2011,wuClosedformSolutionsGeneralized2012,wuExplicitSolutionsMatrix2013},
they have attracted many researchers’ interests. CCME originally 
starts as the following matrix equation
\cite{bevisConsimilarityMatrixEquation1987}: 
\begin{equation} \label{eq.sccsme.invariant}
	A\overline{X}-XB=C.
\end{equation}
CCME \eqref{eq.sccsme.invariant} is also known as standard Sylvester-conjugate matrix equations (SSCME), and it is time-invariant. Solving methods of CCME are shown in Fig. \ref{fig.compare.solve}.

In Fig. \ref{fig.compare.solve}, it is seen that solving time-invariant CCME is mainly based on Wu et al. propose Smith iterative solution
\cite{wuClosedformSolutionsNonhomogeneous2009,wuCompleteSolutionSylvesterpolynomialconjugate2011,wuParametricSolutionsSylvesterconjugate2011},
gradient-based iterative solution 
\cite{dingHierarchicalGradientbasedIdentification2005,dingHierarchicalLeastSquares2005,wuIterativeAlgorithmsSolving2011,wangModifiedGradientBased2012,bayoumiAcceleratedGradientbasedIterative2016,ZHANG20177585} 
and finite iterative solution
\cite{wuFiniteIterativeAlgorithms2011,wuFiniteIterativeAlgorithms2011a},
etc. 
However, in real scientific scenarios, many equations are based on fixed matrix elements at a fixed time. In the long run, the matrix equations are time-variant, like converting SSCME \eqref{eq.sccsme.invariant} to time-variant standard Sylvester-conjugate matrix equations (TVSSCME) \eqref{eq.sccsme.invariant.tranvariant} as below:
\begin{equation} \label{eq.sccsme.invariant.tranvariant}
	A(\tau)\overline{X}(\tau)-X(\tau)B(\tau)=C(\tau),
\end{equation}
where $\tau \ge 0$ denotes the real-time. In known studies, TVSSCME \eqref{eq.sccsme.invariant.tranvariant} lacks proposed theories of solution.
Zhang et al. illustrate that the solution of time-variant matrix equations is also applicable to the solution of time-invariant matrix equations
\cite{uhligZhangNeuralNetworks2024,zhangMoreNewtonIterations2010,jinZeroingNeuralNetworks2017}, i.e., time-variant matrices are extensions of time-invariant matrices. So it is necessary to study the solution of such time-variant matrix equations.

Artificial neural networks
\cite{uhligZhangNeuralNetworks2024,Luo2024NewZN,Gerontitis2022AFO,xiaoDesignVerificationRobotic2018,zhangNewVaryingParameterRecurrent2018,Jiang2023ANZ,liFiniteTimeConvergentNoiseRejection2020,jiangModelingReasoningApplication2022,longFinitetimeStabilizationComplexvalued2022,gerontitis_robust_2022}
play an important role in solving matrix equations. For example, 
gradient neural network (GNN)
\cite{zhangGNNModelTimeVarying2024}
is proposed to solve standard Sylvester matrix equations
\cite{xiaoConvergenceenhancedGradientNeural2017},
Moore-Penrose generalized inverse 
\cite{lvImprovedGradientNeural2019}, etc. It is known that GNN's solution depends on the gradient descent method and the derivation of error pairs defined by least squares
\cite{zhangSimulationComparisonZhang2008}.
However, it is unable to deal with matrix equations that have $X^{\mathrm{T}}$, $X^{\mathrm{H}}$ or $\overline{X}$ directly. Especially encountered matrix equations containing $X$ and $\overline{X}$, the gradient under the derivation requires special treatment
such as partial derivatives and real representations, etc. Therefore, zeroing neural network (ZNN)
\cite{uhligZhangNeuralNetworks2024,jinZeroingNeuralNetworks2017,Xiao2021AnAZ} proposed by Zhang et al. is considered to be used. ZNN is based on the time-variant equation-based neural network solution, in the time-variant matrix equations
\cite{xiaoNewRecurrentNeural2018,lvImprovedZhangNeural2019,xiaoComputingTimeVaryingQuadratic2019,xiaoNewNoisetolerantPredefinedtime2019,xiaoDesignComprehensiveAnalysis2020,xiaoNewErrorFunction2020} 
have an absolute advantage
with comparing GNN, which is directly defined in terms of the error function, avoiding the GNN's gradient derivation and the defect of lagging error. And zeroing neural dynamics (ZND)
\cite{lvWsbpFunctionActivated2018,liaoDifferentZFsDifferent2014,WU2022391,10355921,WU202344},
by the development of the real field to the complex field 
\cite{liNonlinearlyActivatedNeural2014,liaoDifferentComplexZFs2014,zhangContinuousDiscreteZeroing2021,xiaoDesignAnalysisNew2019,xiaoNovelRecurrentNeural2019,fuZNDZeaDModelsTheoretics2021,xiaoPredefinedtimeAntinoiseVaryingparameter2023}, 
solving many time-variant matrix equations with $X^{\mathrm{T}}(\tau)$ or $X^{\mathrm{H}}(\tau)$, where $\tau \ge 0$ denotes the real-time.
But based on TVSSCME \eqref{eq.sccsme.invariant.tranvariant} with $\overline{X}(\tau)$, there is no unified and systematic theory. So the solution of TVSSCME \eqref{eq.sccsme.invariant.tranvariant} study not only complements the solution of SSCME \eqref{eq.sccsme.invariant} but also improves related theory of solving time-variant CCME by ZNN.

The rest of the paper is organized as follows.
Section 2 provides the definition of TVSSCME and additional knowledge for solving this class of equations. 
Section 3 proposes ZND solution model in the complex field and real field. 
Section 4 gives simulations to verify the validity of the models and compare the advantages and disadvantages of each model. 
Section 5 concludes the paper and suggests future directions. Before starting the next section, the main contributions of this paper are listed as follows.

\begin{itemize}
	\item[(1)]Based on ZND, Con-CZND1 model and Con-CZND2 model are proposed to solve TVSSCME for the first time in known studies. 
	
	\item[(2)]By redefining the vectorization and Kronecker product, the convergence and the effectiveness of the proposition is proved theoretically.  
 
	\item[(3)]Three numerical comparison experiments are done to highlight the 
	significance of ZND in the complex field and the complexity of it.
\end{itemize}

\section{Problem Formulation and Preliminaries}
In this section, a present study of TVSSCME
\cite{bevisConsimilarityMatrixEquation1987,bevisMatrixEquationTextbackslashtextbackslashA1988,Wu2017,wuSolutionsMatrixEquations2006}
is researched, while the vectorization and Kronecker product in the complex field are given:
\begin{defi}
According to $XF-A\overline{X}=C$, TVSSCME is formulated as
	\begin{align} \label{eq.sccsme.variant}
		X(\tau)F(\tau)-A(\tau)\overline{X}(\tau)=C(\tau),
	\end{align}
	where $F(\tau)\in
	\mathbb{C}^{n\times n}$, $A(\tau)\in
	\mathbb{C}^{m\times m}$, $C(\tau)\in
	\mathbb{C}^{m\times n}$ are known time-variant matrices,
	and $\tau \ge 0$ denotes the real-time.
	In addition,
	$X(\tau)\in
	\mathbb{C}^{m\times n}$, along with its conjugate version $\overline{X}(\tau)\in
	\mathbb{C}^{m\times n}$, is a time-variant matrix to be computed.
	In this paper, the unique time-variant solution $X(\tau)$ is only considered here.
	Then $X^*(\tau)\in
	\mathbb{C}^{m\times n}$ is defined as the exact time-variant matrix-form solution of
	TVSSCME \eqref{eq.sccsme.variant}, where the time-variant matrix in the complex field can be described as
\end{defi}
\begin{defi}
Where $\tau \ge 0$ denotes the real-time, the time-variant matrix in the complex field is defined as
\begin{align}\label{eq.define.complexmatrix}
	M(\tau)
	&=\begin{bmatrix}
		\tilde{m}_{11}(\tau)  &\cdots   &\tilde{m}_{1q}(\tau) \\
		\vdots &\ddots   &\vdots  \\
		\tilde{m}_{p1}(\tau)&\cdots  &\tilde{m}_{pq}(\tau)
	\end{bmatrix}
	\notag\\
	&=\begin{bmatrix}
		m_{\mathrm{r},11}(\tau)+\mathrm{i}m_{\mathrm{i},11}(\tau)  &\cdots   &m_{\mathrm{r},1q}(\tau)+\mathrm{i}m_{\mathrm{i},1q}(\tau) \\
		\vdots &\ddots   &\vdots  \\
		m_{\mathrm{r},p1}(\tau)+\mathrm{i}m_{\mathrm{i},p1}(\tau)&\cdots  &m_{\mathrm{r},pq}(\tau)+\mathrm{i}m_{\mathrm{i},pq}(\tau)
	\end{bmatrix}
	\notag\\
	&=\begin{bmatrix}
	m_{\mathrm{r},11}(\tau)  &\cdots   &m_{\mathrm{r},1q}(\tau) \\
		\vdots &\ddots   &\vdots  \\
		m_{\mathrm{r},p1}(\tau)&\cdots  &m_{\mathrm{r},pq}(\tau)
	\end{bmatrix}+\mathrm{i}
	\begin{bmatrix}
		m_{\mathrm{i},11}(\tau)  &\cdots   &m_{\mathrm{i},1q}(\tau) \\
		\vdots &\ddots   &\vdots  \\
		m_{\mathrm{i},p1}(\tau)&\cdots  &m_{\mathrm{i},pq}(\tau)
	\end{bmatrix}
	\notag\\
	&=M_{\mathrm{r}}(\tau)+\mathrm{i}M_{\mathrm{i}}(\tau),
	\end{align}
	where $M(\tau)\in
	\mathbb{C}^{p\times q}$ is any complex matrix, $\mathrm{i}$ is an imaginary unit. For simplicity, this paper uses $\tilde{m}_{st}(\tau)$ to represent the complex elements of the matrix, where $s\in
	\mathbb{I}[1,p]$, $t\in
	\mathbb{I}[1,q]$, $\mathbb{I}[m,n]$ means the set of integers from m to n, same as below. Because its elements are complex numbers, they can be expressed as $\tilde{m}_{st}(\tau)=m_{\mathrm{r},st}(\tau)+\mathrm{i}m_{\mathrm{i},st}(\tau)$.
	Thus, $M_{\mathrm{r}}(\tau)\in
	\mathbb{R}^{p\times q}$ is the real coefficient matrix of $M(\tau)$, where $m_{\mathrm{r},st}(\tau)$ is a real coefficient; $M_{\mathrm{i}}(\tau)\in
	\mathbb{R}^{p\times q}$ is the imaginary coefficient matrix of $M(\tau)$, where $m_{\mathrm{i},st}(\tau)$ is a imaginary coefficient. The conjugate matrix corresponding to $M(\tau)$ is $\overline{M}(\tau)=M_{\mathrm{r}}(\tau)-\mathrm{i}M_{\mathrm{i}}(\tau)$, where $\overline{M}(\tau)\in
	\mathbb{C}^{p\times q}$.
\end{defi}
\begin{defi}
	According to \eqref{eq.define.complexmatrix}, where $\tau \ge 0$ denotes the real-time, the time-variant matrix single transpose in the complex field is defined as
	\begin{align} \label{eq.define.complexsingletranspose}
		M^{\mathrm{T}}(\tau)
	&=\begin{bmatrix}
		\tilde{m}_{11}(\tau)  &\cdots   &\tilde{m}_{p1}(\tau) \\
		\vdots &\ddots   &\vdots  \\
		\tilde{m}_{1q}(\tau)&\cdots  &\tilde{m}_{pq}(\tau)
	\end{bmatrix}
	\notag\\
	&=\begin{bmatrix}
		m_{\mathrm{r},11}(\tau)+\mathrm{i}m_{\mathrm{i},11}(\tau)  &\cdots   &m_{\mathrm{r},p1}(\tau)+\mathrm{i}m_{\mathrm{i},p1}(\tau) \\
		\vdots &\ddots   &\vdots  \\
		m_{\mathrm{r},1q}(\tau)+\mathrm{i}m_{\mathrm{i},1q}(\tau)&\cdots  &m_{\mathrm{r},pq}(\tau)+\mathrm{i}m_{\mathrm{i},pq}(\tau)
	\end{bmatrix}
		\notag\\
		&=\overline{M^{\mathrm{H}}}(\tau),
	\end{align}
	where $M^{\mathrm{T}}(\tau)\in
	\mathbb{C}^{q\times p}$ is any complex matrix.
	However, since matrices are defined in the complex field and operations are mostly based on conjugate transpositions. In order to unify above operations, $\overline{M^{\mathrm{H}}}(\tau)\in
	\mathbb{C}^{q\times p}$ is used to denote the simple transposition. Because of \eqref{eq.define.complexsingletranspose}, the real field matrices apply to this definition as well.
\end{defi}
\begin{defi}
	Where $F(\tau)\in
	\mathbb{R}^{n\times n}$, $A(\tau)\in
	\mathbb{R}^{m\times m}$, $C(\tau)\in
	\mathbb{R}^{m\times n}$ are known time-variant matrices,
	$X(\tau)\in
	\mathbb{R}^{m\times n}$ is a time-variant matrix to be computed,
	and $\tau \ge 0$ denotes the real-time, over the real field, the definition of the time-variant standard Sylvester matrix equations (TVSSME) is given as
	\begin{align} \label{eq.ssme.variant}
		X(\tau)F(\tau)-A(\tau)X(\tau)=C(\tau).
	\end{align}
	
	TVSSCME \eqref{eq.sccsme.variant} 
	is an extensive version of TVSSME
	\eqref{eq.ssme.variant}
	\cite{wuSolutionsMatrixEquations2006}.
	When the coefficients of the imaginary matrices of TVSSCME \eqref{eq.sccsme.variant}
	are all zero, it degenerates into 
	TVSSME \eqref{eq.ssme.variant} 
	in the real field.
\end{defi}
\begin{defi}
	According to 
	\eqref{eq.define.complexmatrix}, where $\tau \ge 0$ denotes the real-time, 
	$\mathrm{vec}(M\\(\tau))$
	is defined as follows.
	\begin{align}\label{eq.define.vectorization}
	\mathrm{vec}(M(\tau))
	&=\begin{bmatrix}
		\tilde{m}_{11}(\tau)\\
		\vdots\\
		\tilde{m}_{p1}(\tau)\\
		\tilde{m}_{12}(\tau)\\
		\vdots\\
		\tilde{m}_{p2}(\tau)\\
		\vdots\\
		\vdots\\
		\tilde{m}_{pq}(\tau)
    \end{bmatrix}  
    =\begin{bmatrix}
	m_{\mathrm{r},11}(\tau)+\mathrm{i}m_{\mathrm{i},11}(\tau)\\
	\vdots\\
	m_{\mathrm{r},p1}(\tau)+\mathrm{i}m_{\mathrm{i},p1}(\tau)\\
	m_{\mathrm{r},12}(\tau)+\mathrm{i}m_{\mathrm{i},12}(\tau)\\
	\vdots\\
	m_{\mathrm{r},p2}(\tau)+\mathrm{i}m_{\mathrm{i},p2}(\tau)\\
	\vdots\\
	\vdots\\
	m_{\mathrm{r},pq}(\tau)+\mathrm{i}m_{\mathrm{i},pq}(\tau)
	\end{bmatrix}
	\notag\\
	&=\begin{bmatrix}
		m_{\mathrm{r},11}(\tau)\\
		\vdots\\
		m_{\mathrm{r},p1}(\tau)\\
		m_{\mathrm{r},12}(\tau)\\
		\vdots\\
		m_{\mathrm{r},p2}(\tau)\\
		\vdots\\
		\vdots\\
		m_{\mathrm{r},pq}(\tau)
	\end{bmatrix}+\mathrm{i}
	\begin{bmatrix}
		m_{\mathrm{i},11}(\tau)\\
		\vdots\\
		m_{\mathrm{i},p1}(\tau)\\
		m_{\mathrm{i},12}(\tau)\\
		\vdots\\
		m_{\mathrm{i},p2}(\tau)\\
		\vdots\\
		\vdots\\
		m_{\mathrm{i},pq}(\tau)
	\end{bmatrix}
	=\mathrm{vec}(M_{\mathrm{r}}(\tau))+\mathrm{i}\mathrm{vec}(M_{\mathrm{i}}(\tau)),
	\end{align}
	where $\mathrm{vec}(M(\tau))\in
	\mathbb{C}^{pq\times 1}$ is any complex vector, $\mathrm{vec}(M_{\mathrm{r}}(\tau))\in
	\mathbb{R}^{pq\times 1}$ is any real coefficient vector of $\mathrm{vec}(M(\tau))$,
	$\mathrm{vec}(M_{\mathrm{i}}(\tau))\in
	\mathbb{R}^{pq\times 1}$ is any imaginary coefficient vector of $\mathrm{vec}(M(\tau))$.
\end{defi}
\begin{defi}
	According to 
	\eqref{eq.define.complexmatrix}, where $A(\tau)\in
	\mathbb{C}^{m\times n}$, $B(\tau)\in
	\mathbb{C}^{s\times t}$ are time-variant matrices
	and $\tau \ge 0$ denotes the real-time, the Kronecker product between them in the complex field is defined as follows.
	\begin{align} \label{eq.define.kroneckerproduct}
	A(\tau)\otimes B(\tau)
	&=\begin{bmatrix}
		\tilde{a}_{11}(\tau)B(\tau)&\tilde{a}_{12}(\tau)B(\tau)  &\cdots  &\tilde{a}_{1n}(\tau)B(\tau) \\
		\tilde{a}_{21}(\tau)B(\tau)&\tilde{a}_{22}(\tau)B(\tau)  &\cdots  &\tilde{a}_{2n}(\tau)B(\tau) \\
		\vdots &\vdots  &\ddots   &\vdots \\
		\tilde{a}_{m1}(\tau)B(\tau)&\tilde{a}_{m2}(\tau)B(\tau)  &\cdots  &\tilde{a}_{mn}(\tau)B(\tau)
	\end{bmatrix},
	\end{align}
	where $\eqref{eq.define.kroneckerproduct}\in
	\mathbb{C}^{ms\times nt}$. 
	Thus, the equation for the vectorization of $A(\tau)X(\tau)B(\tau)$ product in the complex field can be obtained in Theorem \ref{thm.complexkroneckerproductvectorization}.
\end{defi}
\begin{thm}\label{thm.complexkroneckerproductvectorization}
	Where $A(\tau)\in
	\mathbb{C}^{m\times n}$, $B(\tau)\in
	\mathbb{C}^{s\times t}$, $X(\tau)\in
	\mathbb{C}^{n\times s}$, are time-variant matrices, and $\tau \ge 0$ denotes the real-time,
	the following equation can be obtained:
	\begin{align}\label{eq.infer.complexkroneckerproductvectorization} 
	\mathrm{vec}(A(\tau)X(\tau)B(\tau))=(\overline{B^{\mathrm{H}}}(\tau)\otimes A(\tau))\mathrm{vec}(X(\tau)).	
	\end{align}
	\begin{pot}
		
		For any complex matrix
		$D(\tau)\in
		\mathbb{C}^{s\times t}$, it is represented in column chunks as
		\begin{align}\label{eq.divide.columns} 
		D(\tau)
		&=\begin{bmatrix}
		\mathbf{\tilde{d}}_{1}(\tau)&\mathbf{\tilde{d}}_{2}(\tau)  &\cdots   &\mathbf{\tilde{d}}_{t}(\tau)
		\end{bmatrix},	
		\end{align} 
		where $\mathbf{\tilde{d}}_{j}(\tau)\in
		\mathbb{C}^{s}$, $j\in
		\mathbb{I}[1,t]$. On this basis, let
		\begin{align}\label{eq.divide.rows}
		\mathbf{\tilde{d}}_{j}(\tau)=
		\begin{bmatrix}
		\tilde{d}_{1j}(\tau)\\
		\tilde{d}_{2j}(\tau)\\
		\vdots \\
		\tilde{d}_{sj}(\tau)
		\end{bmatrix},
        \end{align}
        according to \eqref{eq.define.complexmatrix},
        \eqref{eq.define.complexsingletranspose},
        \eqref{eq.define.vectorization}, \eqref{eq.define.kroneckerproduct}, \eqref{eq.divide.columns}, and \eqref{eq.divide.rows}, the following equation can be obtained:
        \begin{align}\label{eq.explain.kroneckerproductvectorization} 
        	&\mathrm{vec}(A(\tau)X(\tau)B(\tau))
        	\notag\\
        	=&\mathrm{vec}(\begin{bmatrix}
        		A(\tau)X(\tau)\mathbf{\tilde{b}}_{1}(\tau)&A(\tau)X(\tau)\mathbf{\tilde{b}}_{2}(\tau)&  \cdots &A(\tau)X(\tau)\mathbf{\tilde{b}}_{t}(\tau)
        	\end{bmatrix})
        	\notag\\
        	=&\begin{bmatrix}
        	A(\tau)X(\tau)\mathbf{\tilde{b}}_{1}(\tau)	\\
        	A(\tau)X(\tau)\mathbf{\tilde{b}}_{2}(\tau)	\\
        	\vdots	\\
        	A(\tau)X(\tau)\mathbf{\tilde{b}}_{t}(\tau)	
        	\end{bmatrix}
        	\notag\\
        	=&\begin{bmatrix}
        		\left(A(\tau)\begin{bmatrix}
        			\mathbf{\tilde{x}}_{1}(\tau)&\mathbf{\tilde{x}}_{2}(\tau)  &\cdots   &\mathbf{\tilde{x}}_{s}(\tau)
        		\end{bmatrix}\begin{bmatrix}
        		\tilde{b}_{11}(\tau)\\
        		\tilde{b}_{21}(\tau)\\
        		\vdots \\
        		\tilde{b}_{s1}(\tau)
        		\end{bmatrix}\right )	\\
        		\\
        		\left(A(\tau)\begin{bmatrix}
\mathbf{\tilde{x}}_{1}(\tau)&\mathbf{\tilde{x}}_{2}(\tau)  &\cdots   &\mathbf{\tilde{x}}_{s}(\tau)
\end{bmatrix}\begin{bmatrix}
\tilde{b}_{12}(\tau)\\
\tilde{b}_{22}(\tau)\\
\vdots \\
\tilde{b}_{s2}(\tau)
\end{bmatrix}\right )	\\\\
        		\vdots	\\\\
        		\left(A(\tau)\begin{bmatrix}
\mathbf{\tilde{x}}_{1}(\tau)&\mathbf{\tilde{x}}_{2}(\tau)  &\cdots   &\mathbf{\tilde{x}}_{s}(\tau)
\end{bmatrix}\begin{bmatrix}
\tilde{b}_{1t}(\tau)\\
\tilde{b}_{2t}(\tau)\\
\vdots \\
\tilde{b}_{st}(\tau)
\end{bmatrix}\right )	
        	\end{bmatrix}
        \notag\\
        =&\begin{bmatrix}
        	\tilde{b}_{11}(\tau)A(\tau)\mathbf{\tilde{x}}_{1}(\tau)+\tilde{b}_{21}(\tau)A(\tau)\mathbf{\tilde{x}}_{2}(\tau)+\cdots+\tilde{b}_{s1}(\tau)A(\tau)\mathbf{\tilde{x}}_{s}(\tau)\\
        	\tilde{b}_{12}(\tau)A(\tau)\mathbf{\tilde{x}}_{1}(\tau)+\tilde{b}_{22}(\tau)A(\tau)\mathbf{\tilde{x}}_{2}(\tau)+\cdots+\tilde{b}_{s2}(\tau)A(\tau)\mathbf{\tilde{x}}_{s}(\tau)\\
        	\cdots\\
        	\tilde{b}_{1t}(\tau)A(\tau)\mathbf{\tilde{x}}_{1}(\tau)+\tilde{b}_{2t}(\tau)A(\tau)\mathbf{\tilde{x}}_{2}(\tau)+\cdots+\tilde{b}_{st}(\tau)A(\tau)\mathbf{\tilde{x}}_{s}(\tau)
        \end{bmatrix}
        \notag\\
        =&\begin{bmatrix}
        	\tilde{b}_{11}(\tau)A(\tau)&\tilde{b}_{21}(\tau)A(\tau)  &\cdots  &\tilde{b}_{s1}(\tau)A(\tau) \\
        	\tilde{b}_{12}(\tau)A(\tau)&\tilde{b}_{22}(\tau)A(\tau)  &\cdots  &\tilde{b}_{s2}(\tau)A(\tau) \\
        	\vdots&\vdots  &\ddots   &\vdots \\
        	\tilde{b}_{1t}(\tau)A(\tau)&\tilde{b}_{2t}(\tau)A(\tau)  &\cdots  &\tilde{b}_{st}(\tau)A(\tau) \\
        \end{bmatrix}
        \begin{bmatrix}
        	\mathbf{\tilde{x}}_{1}(\tau)\\
        	\mathbf{\tilde{x}}_{2}(\tau)\\
        	\vdots\\
        	\mathbf{\tilde{x}}_{s}(\tau)
        \end{bmatrix}
        \notag\\
         =& \left (\begin{bmatrix}
        	\tilde{b}_{11}(\tau)&\tilde{b}_{21}(\tau) &\cdots  &\tilde{b}_{s1}(\tau) \\
        	\tilde{b}_{12}(\tau)&\tilde{b}_{22}(\tau)  &\cdots  &\tilde{b}_{s2}(\tau) \\
        	\vdots&\vdots  &\ddots   &\vdots \\
        	\tilde{b}_{1t}(\tau)&\tilde{b}_{2t}(\tau)  &\cdots  &\tilde{b}_{st}(\tau) \\
        \end{bmatrix}\otimes A(\tau) \right ) 
        \mathrm{vec}(X(\tau))
        \notag\\
        =&(\overline{B^{\mathrm{H}}}(\tau)\otimes A(\tau))\mathrm{vec}(X(\tau)).	
        \end{align}
        
		The proof is thus completed.\qed
	\end{pot}
\end{thm}
\begin{corollary}
When the imaginary matrix coefficients of $A(\tau)
$, $B(\tau)
$, $X(\tau)$ are all zero,
i.e. $A(\tau)\in
\mathbb{R}^{m\times n}$, $B(\tau)\in
\mathbb{R}^{s\times t}$, $X(\tau)\in
\mathbb{R}^{n\times s}$, \eqref{eq.infer.complexkroneckerproductvectorization} degenerates into the time-variant equivalence under the real field likes: \begin{align}\label{eq.infer.realkroneckerproductvectorization} 
	\mathrm{vec}(A(\tau)X(\tau)B(\tau))=(B^{\mathrm{T}}(\tau)\otimes A(\tau))\mathrm{vec}(X(\tau)).	
\end{align}

So taking the time-variant equation under the real field
\eqref{eq.infer.realkroneckerproductvectorization} is a special case of the complex field \eqref{eq.infer.complexkroneckerproductvectorization}.
\end{corollary}

\section{Models, Algorithm and Analyses}
In this section, based on previous basics, two models for dealing with TVSSCME \eqref{eq.sccsme.variant} are proposed. The first model involves directly addressing it on the complex field ZND \cite{fuZNDZeaDModelsTheoretics2021}. 
Then, it can be transformed in the real field, which is called Con-CZND1 discussed in Section 3.1. The second model first separates the real and imaginary coefficients matrices. Then, it is substituted into the real field ZND \cite{8809787} to solve, which is called Con-CZND2 discussed in Section 3.2.  

\subsection{Con-CZND1 model}
Based on \eqref{eq.infer.complexkroneckerproductvectorization}, the complex field ZND 
\cite{fuZNDZeaDModelsTheoretics2021}
is proposed to solve TVSSCME \eqref{eq.sccsme.variant}.

Firstly, the error function is defined as follows.
\begin{align} \label{eq.define.errconcznd1}
	E_{\mathrm{M1}}(\tau)=X(\tau)F(\tau)-A(\tau)\overline{X}(\tau)-C(\tau),
\end{align}
	where $E_{\mathrm{M1}}(\tau)\in
	\mathbb{C}^{m\times n}$. Next, the formula in the complex field ZND is proposed to make all elements of \eqref{eq.define.errconcznd1} 
	converge to zero, which is obtained as
\begin{align} \label{eq.deduce.errconcznd1}
	\frac{\partial E_{\mathrm{M1}}(\tau)}{\partial \tau} =-\gamma \Phi \left ( E_{\mathrm{M1}}(\tau) \right ),
\end{align}
where $\gamma\in\mathbb{R^+}$ denotes the regulation parameter controlling the convergence rate \cite{1031938}, and $\Phi \left (\cdot  \right )$ denotes the monotonically increasing odd activation function. For simplicity, a linear activation function is used in this case, and \eqref{eq.deduce.errconcznd1} is simplified to:
\begin{align} \label{eq.infer.linearerrconcznd1}
	\frac{\partial E_{\mathrm{M1}}(\tau)}{\partial \tau} =-\gamma E_{\mathrm{M1}}(\tau).
\end{align}

Then, \eqref{eq.define.errconcznd1} is substituted into \eqref{eq.infer.linearerrconcznd1} to obtain \eqref{eq.join.linearerrconcznd1}:
\begin{align} \label{eq.join.linearerrconcznd1}
	\dot{X}(\tau)F(\tau)+X(\tau)\dot{F}(\tau)-\dot{A}(\tau)\overline{X}(\tau)-A(\tau)\dot{\overline{X}}(\tau)-\dot{C}(\tau)
	\notag\\
	=-\gamma(X(\tau)F(\tau)-A(\tau)\overline{X}(\tau)-C(\tau)).
\end{align}

Applied \eqref{eq.infer.complexkroneckerproductvectorization} given in Theorem \ref{thm.complexkroneckerproductvectorization}, \eqref{eq.join.linearerrconcznd1} is converted to \eqref{eq.use.complexkroneckerproductvectorization}:
\begin{align} \label{eq.use.complexkroneckerproductvectorization}
	&(\overline{F^{\mathrm{H}}}(\tau)\otimes I_{m})\mathrm{vec}(\dot{X}(\tau))-(\overline{I_{n}^{\mathrm{H}}}\otimes A(\tau))\mathrm{vec}(\dot{\overline{X}}(\tau))
	\notag\\
	=&\mathrm{vec}(\dot{C}(\tau)+\dot{A}(\tau)\overline{X}(\tau)-X(\tau)\dot{F}(\tau))
	\notag\\
	&-\gamma \mathrm{vec} (X(\tau)F(\tau)-A(\tau)\overline{X}(\tau)-C(\tau)).
\end{align}

Then, \eqref{eq.use.complexkroneckerproductvectorization} is further reformulated as
\begin{align}
\label{eq.simplify.complexkroneckerproductvectorization}
&U(\tau)\mathrm{vec}(\dot{X}(\tau))-V(\tau)\mathrm{vec}(\dot{\overline{X}}(\tau))
=G(\tau),
\end{align}
where $U(\tau)=(\overline{F^{\mathrm{H}}}(\tau) \otimes I_{m})\in
\mathbb{C}^{nm\times mn}$, $V(\tau)=(\overline{I_{n}^{\mathrm{H}}} \otimes A(\tau))=(I_{n}\otimes A(\tau))\in
\mathbb{C}^{mn\times nm}$,
$G(\tau)=\mathrm{vec}(\dot{C}(\tau)+\dot{A}(\tau)\overline{X}(\tau)-X(\tau)\dot{F}(\tau))
-\gamma \mathrm{vec} (X(\tau)F(\tau)-A(\tau)\overline{X}(\tau)-C(\tau))\in
\mathbb{C}^{mn\times 1}$. Based on the linearity of the derivative as well as \eqref{eq.define.complexmatrix} and \eqref{eq.define.vectorization}, \eqref{eq.simplify.complexkroneckerproductvectorization} can be written in the form of the following real-only matrix operation:
\begin{align}\label{eq.divide.complexkroneckerproductvectorization}
\begin{bmatrix}
U_{\mathrm{r}}(\tau)-V_{\mathrm{r}}(\tau)	&-(U_{\mathrm{i}}(\tau)+V_{\mathrm{i}}(\tau)) \\
U_{\mathrm{i}}(\tau)-V_{\mathrm{i}}(\tau)	&U_{\mathrm{r}}(\tau)+V_{\mathrm{r}}(\tau)
\end{bmatrix}
\begin{bmatrix}
\dot{Z}_{\mathrm{r}}(\tau)\\
\dot{Z}_{\mathrm{i}}(\tau)	
\end{bmatrix}
=\begin{bmatrix}
G_{\mathrm{r}}(\tau)\\
G_{\mathrm{i}}(\tau)	
\end{bmatrix},
\end{align}
where $Z(\tau)=\mathrm{vec}(X(\tau))\in
\mathbb{C}^{mn\times 1}$, $\dot{Z}(\tau)=\mathrm{vec}(\dot{X}(\tau))\in
\mathbb{C}^{mn\times 1}$. To simplify, let $W_{\mathrm{M1}}(\tau)=\left [U_{\mathrm{r}}(\tau)-V_{\mathrm{r}}(\tau), -(U_{\mathrm{i}}(\tau)+V_{\mathrm{i}}(\tau));
U_{\mathrm{i}}(\tau)-V_{\mathrm{i}}(\tau),	U_{\mathrm{r}}(\tau)+V_{\mathrm{r}}(\tau)\right ]
\in\mathbb{R}^{2mn\times 2mn}$, $\dot{X}_{\mathrm{M1}}(\tau)= \left [\dot{Z}_{\mathrm{r}}(\tau);
\dot{Z}_{\mathrm{i}}(\tau)\right ]\in
\mathbb{R}^{2mn\times 1}$, $B_{\mathrm{M1}}(\tau)= \left [G_{\mathrm{r}}(\tau);
G_{\mathrm{i}}(\tau)\right ]\in
\mathbb{R}^{2mn\times 1}$. The final solution model Con-CZND1 is obtained:
\begin{align} \label{eq.solve.linearerrconcznd1}
	\dot{X}_{\mathrm{M1}}(\tau) =W^{+}_{\mathrm{M1}}(\tau)B_{\mathrm{M1}}(\tau),
\end{align}
where $W^{+}_{\mathrm{M1}}(\tau)$ is the pseudo-inverse matrix of $W_{\mathrm{M1}}(\tau)$.
\begin{thm}
	Given differentiable time-variant matrices $F(\tau)\in
	\mathbb{C}^{n\times n}$, $A(\tau)\\\in
	\mathbb{C}^{m\times m}$, and $C(\tau)\in
	\mathbb{C}^{m\times n}$, if TVSSCME \eqref{eq.sccsme.variant} only has one theoretical time-variant solution $X^*(\tau)\in
	\mathbb{C}^{m\times n}$, then each solving element of \eqref{eq.solve.linearerrconcznd1} converges to the corresponding theoretical time-variant solving elements.
	\begin{pot}
		According to 
		\cite{liNonlinearlyActivatedNeural2014},
		based on \eqref{eq.define.complexmatrix}, 
		\eqref{eq.deduce.errconcznd1} can be derived in its equivalent form as \begin{align}\label{eq.eqdeduce.errconcznd1}
			\dot{E}_{\mathrm{M1}}(\tau)=-\gamma \Phi \left ( E_{\mathrm{M1}_{\mathrm{r}}}(\tau)+\mathrm{i}E_{\mathrm{M1}_{\mathrm{i}}}(\tau) \right ),	
		\end{align}
		where $\dot{E}_{\mathrm{M1}}(\tau)\in
		\mathbb{C}^{m\times n}$, with its elements $\dot{\tilde{e}}_{\mathrm{M1}_{st}}(\tau)\in
		\mathbb{C}$;
		$E_{\mathrm{M1}_{\mathrm{r}}}(\tau)\in
		\mathbb{R}^{m\times n}$, with its elements $e_{\mathrm{M1}_{\mathrm{r},st}}(\tau)\in
		\mathbb{R}$;
		$E_{\mathrm{M1}_{\mathrm{i}}}(\tau)\in
		\mathbb{R}^{m\times n}$
		, with its elements $e_{\mathrm{M1}_{\mathrm{i},st}}(\tau)
		\in
		\mathbb{R}$; $s\in
		\mathbb{I}[1,m]$, $t\in
		\mathbb{I}[1,n]$.
		
		Then, using the linearity of the derivative, combined with the completeness of the set of complex numbers and the closure of complex arithmetic, \eqref{eq.eqdeduce.errconcznd1} is split into the following two equations:
		\begin{align}\label{eq.eqdeduce1.errconcznd1}
		\dot{E}_{\mathrm{M1}_{\mathrm{r}}}(\tau)=-\gamma \Phi \left ( E_{\mathrm{M1}_{\mathrm{r}}}(\tau) \right ),	
		\end{align}
		\begin{align}\label{eq.eqdeduce2.errconcznd1}
		\dot{E}_{\mathrm{M1}_{\mathrm{i}}}(\tau)=-\gamma \Phi \left ( E_{\mathrm{M1}_{\mathrm{i}}}(\tau) \right ),	
		\end{align}
		where $\dot{E}_{\mathrm{M1}_{\mathrm{r}}}(\tau)\in
		\mathbb{R}^{m\times n}$, with its elements $\dot{e}_{\mathrm{M1}_{\mathrm{r},st}}(\tau)\in
		\mathbb{R}$, that $\dot{e}_{\mathrm{M1}_{\mathrm{r},st}}(\tau)=-\gamma \Phi \left ( e_{\mathrm{M1}_{\mathrm{r},st}}(\tau) \right )$;
		$\dot{E}_{\mathrm{M1}_{\mathrm{i}}}(\tau)\in
		\mathbb{R}^{m\times n}$
		, with its elements $\dot{e}_{\mathrm{M1}_{\mathrm{i},st}}(\tau)
		\in
		\mathbb{R}$, that $\dot{e}_{\mathrm{M1}_{\mathrm{i},st}}(\tau)=-\gamma \Phi \left ( e_{\mathrm{M1}_{\mathrm{i},st}}(\tau) \right )$.
		
		As shown above, the time-variant systems represented by both of the \eqref{eq.eqdeduce1.errconcznd1} and \eqref{eq.eqdeduce2.errconcznd1} are actually design formulas for real field ZND
		\cite{liNonlinearlyActivatedNeural2014}. According to the stability of real field ZND, both the above \eqref{eq.eqdeduce1.errconcznd1} and \eqref{eq.eqdeduce2.errconcznd1} are stable if they have the monotonically increasing odd activation function. Then, again according to Lyapunov theory, two Lyapunov functions are designed: $V_{\mathrm{M1}_{\mathrm{r}}}(\tau)=\frac{1}{2} e^2_{\mathrm{M1}_{\mathrm{r},st}}(\tau)$ and $V_{\mathrm{M1}_{\mathrm{i}}}(\tau)=\frac{1}{2} e^2_{\mathrm{M1}_{\mathrm{i},st}}(\tau)$.
		Then two functions are derived and each is inferred as follows.
		\begin{align}\label{eq.eqdeduce1.errconcznd1elements}
			\dot{V}_{\mathrm{M1}_{\mathrm{r}}}(\tau)= e_{\mathrm{M1}_{\mathrm{r},st}}(\tau)\dot{e}_{\mathrm{M1}_{\mathrm{r},st}}(\tau)=-\gamma e_{\mathrm{M1}_{\mathrm{r},st}}(\tau)\Phi \left ( e_{\mathrm{M1}_{\mathrm{r},st}}(\tau) \right )\le 0,	
		\end{align}
		\begin{align}\label{eq.eqdeduce2.errconcznd1elements}
			\dot{V}_{\mathrm{M1}_{\mathrm{i}}}(\tau)= e_{\mathrm{M1}_{\mathrm{i},st}}(\tau)\dot{e}_{\mathrm{M1}_{\mathrm{i},st}}(\tau)=-\gamma e_{\mathrm{M1}_{\mathrm{i},st}}(\tau)\Phi \left ( e_{\mathrm{M1}_{\mathrm{i},st}}(\tau) \right )\le 0,	
		\end{align}
		when $\tau\to+\infty$, while 
		\eqref{eq.eqdeduce1.errconcznd1} and \eqref{eq.eqdeduce2.errconcznd1} converge to their equilibrium point, i.e., their parallel elements are all about to converge to zero. Finally, since the above disassembled equations’ elements all converge to zero, and then combine back to \eqref{eq.eqdeduce.errconcznd1}, \eqref{eq.define.errconcznd1}'s elements also converge to zero later. Thus, \eqref{eq.solve.linearerrconcznd1}'s elements finally converge from a random initial state to elements corresponding theoretical time-variant solution.
		
		The proof is thus completed.\qed
	\end{pot}
\end{thm}

\subsection{Con-CZND2 model}
In this part, contrasting to Con-CZND1 \eqref{eq.solve.linearerrconcznd1} model, TVSSCME \eqref{eq.sccsme.variant} does not define the error directly first, but performs an equivalent mapping transformation of the matrix equations before proposing the real field ZND \cite{8809787}.

According to \eqref{eq.define.complexmatrix}, matrix splitting is first performed to obtain the following:
\begin{align}\label{eq.transccsme.variant}
	(X_{\mathrm{r}}(\tau)+\mathrm{i}X_{\mathrm{i}}(\tau))(F_{\mathrm{r}}(\tau)+\mathrm{i}F_{\mathrm{i}}(\tau))-&(A_{\mathrm{r}}(\tau)+\mathrm{i}A_{\mathrm{i}}(\tau))(X_{\mathrm{r}}(\tau)-\mathrm{i}X_{\mathrm{i}}(\tau))
	\notag\\
	&=(C_{\mathrm{r}}(\tau)+\mathrm{i}C_{\mathrm{i}}(\tau)).	
\end{align}

Then, \eqref{eq.transccsme.variant} is  performed separation to obtain \eqref{eq.dividetransccsme.variant} with real-only matrix operation:
\begin{align}\label{eq.dividetransccsme.variant}
	\left\{\begin{matrix}
	X_{\mathrm{r}}(\tau)F_{\mathrm{r}}(\tau)-X_{\mathrm{i}}(\tau)F_{\mathrm{i}}(\tau)-A_{\mathrm{r}}(\tau)X_{\mathrm{r}}(\tau)-A_{\mathrm{i}}(\tau)X_{\mathrm{i}}(\tau)=C_{\mathrm{r}}(\tau),\\
	X_{\mathrm{i}}(\tau)F_{\mathrm{r}}(\tau)+X_{\mathrm{r}}(\tau)F_{\mathrm{i}}(\tau)-A_{\mathrm{i}}(\tau)X_{\mathrm{r}}(\tau)+A_{\mathrm{r}}(\tau)X_{\mathrm{i}}(\tau)=C_{\mathrm{i}}(\tau).	
	\end{matrix}\right.
\end{align}

According to \eqref{eq.infer.realkroneckerproductvectorization}, \eqref{eq.dividetransccsme.variant} is formulated into the following:
\begin{align}\label{eq.veckrondividetransccsme.variant}
	\begin{bmatrix}
		K_{11}(\tau)	&K_{12}(\tau) \\
		K_{21}(\tau)	&K_{22}(\tau)
	\end{bmatrix}
	\begin{bmatrix}
		\mathrm{vec}(X_{\mathrm{r}}(\tau))\\
		\mathrm{vec}(X_{\mathrm{i}}(\tau))	
	\end{bmatrix}
	=\begin{bmatrix}
		\mathrm{vec}(C_{\mathrm{r}}(\tau))\\
		\mathrm{vec}(C_{\mathrm{i}}(\tau))	
	\end{bmatrix},
\end{align}
where $K_{11}(\tau)=(F_{\mathrm{r}}^{\mathrm{T}}(\tau) \otimes I_{m})-(I_{n} \otimes A_{\mathrm{r}}(\tau))\in
\mathbb{R}^{nm\times mn}$, $K_{12}(\tau)=-(F_{\mathrm{i}}^{\mathrm{T}}(\tau) \otimes I_{m}+I_{n} \otimes A_{\mathrm{i}}(\tau))\in
\mathbb{R}^{nm\times mn}$,
$K_{21}(\tau)=(F_{\mathrm{i}}^{\mathrm{T}}(\tau) \otimes I_{m})-(I_{n} \otimes A_{\mathrm{i}}(\tau))\in
\mathbb{R}^{nm\times mn}$,
$K_{22}(\tau)=(F_{\mathrm{r}}^{\mathrm{T}}(\tau) \otimes I_{m}+I_{n}\otimes A_{\mathrm{r}}(\tau))\in
\mathbb{R}^{nm\times mn}$. Then let $W_{\mathrm{M2}}(\tau)=[K_{11}(\tau),K_{12}(\tau);
K_{21}(\tau),K_{22}(\tau)]\in
\mathbb{R}^{2mn\times 2mn}$, $X_{\mathrm{M2}}(\tau)=[\mathrm{vec}(X_{\mathrm{r}}(\tau));
\mathrm{vec}(X_{\mathrm{i}}\\(\tau))]\in\mathbb{R}^{2mn\times 1}$,
$B_{\mathrm{M2}}(\tau)=[\mathrm{vec}(C_{\mathrm{r}}(\tau)); \mathrm{vec}(C_{\mathrm{i}}(\tau))]\in\mathbb{R}^{2mn\times 1}$,
\eqref{eq.time.linearerrconcznd2} is obtained:
\begin{align} \label{eq.time.linearerrconcznd2}
	W_{\mathrm{M2}}(\tau)X_{\mathrm{M2}}(\tau) =B_{\mathrm{M2}}(\tau).
\end{align}

Using real field ZND, the error function is first defined as follows.
\begin{align} \label{eq.define.errconcznd2}
	E_{\mathrm{M2}}(\tau)=W_{\mathrm{M2}}(\tau)X_{\mathrm{M2}}(\tau)-B_{\mathrm{M2}}(\tau),
\end{align}
where $E_{\mathrm{M2}}(\tau)\in\mathbb{R}^{2mn\times 1}$.
Next, the formula under the real field of ZND is proposed to make all elements of \eqref{eq.define.errconcznd2} 
converge to zero, which is obtained as
\begin{align} \label{eq.deduce.errconcznd2}
	\frac{\partial E_{\mathrm{M2}}(\tau)}{\partial \tau} =-\gamma \Phi \left ( E_{\mathrm{M2}}(\tau) \right ).
\end{align}

As in the previous subsection, $\gamma\in\mathbb{R^+}$ denotes the regulation parameter controlling the convergence rate \cite{1031938}, and $\Phi \left (\cdot  \right )$ denotes the monotonically increasing odd activation function. For simplicity, a linear activation function is used in this case, and \eqref{eq.deduce.errconcznd2} is simplified to:
\begin{align} \label{eq.infer.linearerrconcznd2}
	\frac{\partial E_{\mathrm{M2}}(\tau)}{\partial \tau} =-\gamma E_{\mathrm{M2}}(\tau).
\end{align}

Then, \eqref{eq.define.errconcznd2} is substituted into \eqref{eq.infer.linearerrconcznd2} to obtain \eqref{eq.join.linearerrconcznd2}:
\begin{align} \label{eq.join.linearerrconcznd2}
	\dot{W}_{\mathrm{M2}}(\tau)X_{\mathrm{M2}}(\tau)+
	W_{\mathrm{M2}}(\tau)\dot{X}_{\mathrm{M2}}(\tau) -\dot{B}_{\mathrm{M2}}(\tau)
	\notag\\
	=-\gamma(W_{\mathrm{M2}}(\tau)X_{\mathrm{M2}}(\tau) -B_{\mathrm{M2}}(\tau)).
\end{align}

Finally, the final solution model Con-CZND2 is obtained:
\begin{align} \label{eq.solve.linearerrconcznd2}
	\dot{X}_{\mathrm{M2}}(\tau)
	=W^{+}_{\mathrm{M2}}(\tau)(\dot{B}_{\mathrm{M2}}(\tau)-\dot{W}_{\mathrm{M2}}(\tau)X_{\mathrm{M2}}(\tau)
	\notag\\
	-\gamma(W_{\mathrm{M2}}(\tau)X_{\mathrm{M2}}(\tau) -B_{\mathrm{M2}}(\tau))),
\end{align}
where $W^{+}_{\mathrm{M2}}(\tau)$ is the pseudo-inverse matrix of $W_{\mathrm{M2}}(\tau)$. 
\begin{thm}
	Given differentiable time-variant matrices $F(\tau)\in
	\mathbb{C}^{n\times n}$, $A(\tau)\\\in
	\mathbb{C}^{m\times m}$, and $C(\tau)\in
	\mathbb{C}^{m\times n}$, if TVSSCME \eqref{eq.sccsme.variant} only has one theoretical time-variant solution $X^*(\tau)\in
	\mathbb{C}^{m\times n}$, then each solving element of \eqref{eq.solve.linearerrconcznd2} converges to the corresponding theoretical time-variant solving elements.
	\begin{pot}
		
	Since the complex field time-variant matrix equations with imaginary matrix coefficients being all zero can degenerate into the real field time-variant matrix equations, as TVSSCME \eqref{eq.sccsme.variant} degenerates into TVSSME \eqref{eq.ssme.variant}, i.e., \eqref{eq.solve.linearerrconcznd2} is based on \eqref{eq.eqdeduce1.errconcznd1} of Theorem 2, the proof of Theorem 2 is an extension and refinement of the proof of Theorem 3. Then, 
	$E_{\mathrm{M2}}(\tau)\in
	\mathbb{R}^{2mn\times 1}$'s elements $e_{\mathrm{M2}_{s1}}(\tau)\in
	\mathbb{R}$,
	and
	$\dot{E}_{\mathrm{M2}}(\tau)\in
	\mathbb{R}^{2mn\times 1}$'s elements $\dot{e}_{\mathrm{M2}_{s1}}(\tau)\in
	\mathbb{R}$, that $\dot{e}_{\mathrm{M2}_{s1}}(\tau)=-\gamma \Phi \left ( e_{\mathrm{M2}_{s1}}(\tau) \right )$, $s\in
	\mathbb{I}[1,2mn]$. Similar to Theorem 2, Lyapunov function is designed: $V_{\mathrm{M2}}(\tau)=\frac{1}{2} e^2_{\mathrm{M2}_{s1}}(\tau)$, then \eqref{eq.eqdeduce.errconcznd2elements} is obtained:
	\begin{align}\label{eq.eqdeduce.errconcznd2elements}
		\dot{V}_{\mathrm{M2}}(\tau)= e_{\mathrm{M2}_{s1}}(\tau)\dot{e}_{\mathrm{M2}_{s1}}(\tau)=-\gamma e_{\mathrm{M2}_{s1}}(\tau)\Phi \left ( e_{\mathrm{M2}_{s1}}(\tau) \right )\le 0,	
	\end{align}
	when $\tau\to+\infty$, while \eqref{eq.deduce.errconcznd2}'s elements  all converge to zero finally. And \eqref{eq.time.linearerrconcznd2} is equivalent to TVSSCME \eqref{eq.sccsme.variant}.
	
		The proof is thus completed.\qed
	\end{pot}
\end{thm}

\section{Numerical Experimentation and Verification}
In this section, based on the previous propositions, the fundamental distinction between \eqref{eq.solve.linearerrconcznd1} and \eqref{eq.solve.linearerrconcznd2} is below interpreted: Con-CZND1 \eqref{eq.solve.linearerrconcznd1} model is considered as ``error before transformation", with the error term proposed directly by the original complex matrix $E_{\mathrm{M1}}(\tau)\in
\mathbb{C}^{m\times n}$; while Con-CZND2 \eqref{eq.solve.linearerrconcznd2} model is considered as ``transformation before error", where $E_{\mathrm{M2}}(\tau)\in
\mathbb{R}^{2mn\times 1}$ is proposed by the real matrix of the transformation after the complex matrix mapping. Although the error matrix $E_{\mathrm{M2}}(\tau)$ of Con-CZND2 \eqref{eq.solve.linearerrconcznd2} model operates in the real field after transformation, the cost is that its dimension becomes larger. Thus, Con-CZND2 \eqref{eq.solve.linearerrconcznd2} model's overall performance is not as good as that of Con-CZND1 \eqref{eq.solve.linearerrconcznd1} model, which is verified by the following experimental
validation.

The accuracy and stability of Con-CZND1 \eqref{eq.solve.linearerrconcznd1} model and Con-CZND2 \eqref{eq.solve.linearerrconcznd2} model are verified through numerical experiments. Three examples are provided below, where $\mathrm{i}$ denotes the imaginary unit and $s\left (\cdot  \right )$ and $c\left (\cdot  \right )$ denote the trigonometric functions $\sin\left (\cdot  \right )$ and $\cos\left (\cdot  \right )$, respectively.
\begin{example}\label{eq.example1}
	Considering TVSSCME \eqref{eq.sccsme.variant} where the dimension of the square matrix $F(\tau)$ is smaller than that of the square matrix $A(\tau)$ and the number of rows of the matrix $C(\tau)$ and the only solution $X^*(\tau)$ is greater than the number of columns:
\begin{align} 
	F(\tau)
	=\begin{bmatrix}
	600+s(\tau)	&c(\tau) \\
	c(\tau)	&400+s(\tau)
	\end{bmatrix}
	+\mathrm{i}\begin{bmatrix}
	c(\tau)	& s(\tau) \\
	s(\tau) & c(\tau)
	\end{bmatrix}\in
	\mathbb{C}^{2\times 2},
	\notag
\end{align}
\begin{align}
	A(\tau)
	=\begin{bmatrix}
		s(\tau)	&c(\tau)  &1 \\
		-c(\tau)&0  &-s(\tau) \\
		1&0  &1
	\end{bmatrix}
	+\mathrm{i}\begin{bmatrix}
	c(\tau)&-s(\tau)  &0 \\
	s(\tau)&1  &c(\tau) \\
	0&1  &0
	\end{bmatrix}\in
	\mathbb{C}^{3\times 3},
	\notag
\end{align}
\begin{align}
	C(\tau)
	=\begin{bmatrix}
	c_{\mathrm{r},11}(\tau) & c_{\mathrm{r},12}(\tau) \\
	c_{\mathrm{r},21}(\tau) & c_{\mathrm{r},22}(\tau) \\
	c_{\mathrm{r},31}(\tau) & c_{\mathrm{r},32}(\tau)
	\end{bmatrix}
	+\mathrm{i}\begin{bmatrix}
	c_{\mathrm{i},11}(\tau) & c_{\mathrm{i},12}(\tau) \\
	c_{\mathrm{i},21}(\tau) & c_{\mathrm{i},22}(\tau) \\
	c_{\mathrm{i},31}(\tau) & c_{\mathrm{i},32}(\tau)
	\end{bmatrix}\in
	\mathbb{C}^{3\times 2},
	\notag	
\end{align}
where
$c_{\mathrm{r},11}(\tau)=600s(\tau)-4c(\tau)s(\tau)+2c^2(\tau)-1$,
$c_{\mathrm{r},12}(\tau)=s(2\tau)+400c(\tau)-2$, $c_{\mathrm{r},21}(\tau)=s(\tau)-599c(\tau)-c(\tau)s(\tau)+c^2(\tau)$, $c_{\mathrm{r},22}(\tau)=-c(\tau)-399s(\tau)+c(\tau)s(\tau)+c^2(\tau)-1$, 
$c_{\mathrm{r},31}(\tau)=599-s(\tau)+c(\tau)$, $c_{\mathrm{r},32}(\tau)=-c(\tau)+s(\tau)$ 
and 
$c_{\mathrm{i},11}(\tau)=600s(\tau)-2c^2(\tau)+2$, $c_{\mathrm{i},12}(\tau)=s(2\tau)+400c(\tau)+1$, $c_{\mathrm{i},21}(\tau)=-600c(\tau)-3c(\tau)s(\tau)+c^2(\tau)-2$,
$c_{\mathrm{i},22}(\tau)=-400s(\tau)-3c(\tau)s(\tau)-c^2(\tau)-1$,$c_{\mathrm{i},31}(\tau)=s(\tau)+3c(\tau)$,
$c_{\mathrm{i},32}(\tau)=c(\tau)+3s(\tau)+401$.

The only exact solution to this example is
\begin{align}
	X^*(\tau)=\begin{bmatrix}
	s(\tau) & c(\tau) \\
	-c(\tau) & -s(\tau) \\
	1 & 0
\end{bmatrix}+\mathrm{i}\begin{bmatrix}
s(\tau) & c(\tau) \\
-c(\tau) & -s(\tau) \\
0 & 1
\end{bmatrix}\in
\mathbb{C}^{3\times 2}.
\end{align}
\end{example}
\begin{example}\label{eq.example2}
	Considering TVSSCME \eqref{eq.sccsme.variant} where the dimension of the square matrix $F(\tau)$ is greater than that of the square matrix $A(\tau)$ and the number of rows of the matrix $C(\tau)$ and the only solution $X^*(\tau)$ is smaller than the number of columns:
\begin{align} 	
	F(\tau)
	=\begin{bmatrix}
		400+s(\tau)	&c(\tau)  &c(\tau) \\
		c(\tau)&200+s(\tau)  &c(\tau) \\
		c(\tau)&c(\tau)  &300+s(\tau)
	\end{bmatrix}+\mathrm{i}\begin{bmatrix}
		c(\tau)&s(\tau)  &s(\tau) \\
		s(\tau)&c(\tau)  &s(\tau) \\
		s(\tau)&s(\tau)  &c(\tau)
	\end{bmatrix}\in
	\mathbb{C}^{3\times 3},
	\notag
\end{align}
\begin{align}
	A(\tau)
	=\begin{bmatrix}
		s(\tau)	&-c(\tau) \\
		c(\tau)	&-s(\tau)
	\end{bmatrix}+\mathrm{i}\begin{bmatrix}
		c(\tau)	&-s(\tau) \\
		s(\tau) &-c(\tau)
	\end{bmatrix}\in
	\mathbb{C}^{2\times 2},
	\notag
\end{align}
\begin{align}
	C(\tau)
	=\begin{bmatrix}
		c_{\mathrm{r},11}(\tau) & c_{\mathrm{r},12}(\tau) & c_{\mathrm{r},13}(\tau)\\
		c_{\mathrm{r},21}(\tau) & c_{\mathrm{r},22}(\tau) & c_{\mathrm{r},23}(\tau)
	\end{bmatrix}+\mathrm{i}\begin{bmatrix}
		c_{\mathrm{i},11}(\tau) & c_{\mathrm{i},12}(\tau) & c_{\mathrm{i},13}(\tau)\\
		c_{\mathrm{i},21}(\tau) & c_{\mathrm{i},22}(\tau) & c_{\mathrm{i},23}(\tau)
	\end{bmatrix}\in
	\mathbb{C}^{2\times 3},
	\notag	
\end{align}
where
$c_{\mathrm{r},11}(\tau)=c(\tau)+400s(\tau)-4c(\tau)s(\tau)$,
$c_{\mathrm{r},12}(\tau)=-2+201c(\tau)$, $c_{\mathrm{r},13}(\tau)=s(\tau)+2c^2(\tau)+299$, $c_{\mathrm{r},21}(\tau)=-400c(\tau)-s(\tau)-4c(\tau)s(\tau)$,
$c_{\mathrm{r},22}(\tau)=-201s(\tau)-2$, $c_{\mathrm{r},23}(\tau)=-c(\tau)-2c^2(\tau)+1$ 
and 
$c_{\mathrm{i},11}(\tau)=401s(\tau)+2$, $c_{\mathrm{i},12}(\tau)=s(\tau)+4c(\tau)s(\tau)+200c(\tau)$, $c_{\mathrm{i},13}(\tau)=-c(\tau)+2c(\tau)s(\tau)+1$,
$c_{\mathrm{i},21}(\tau)=-399c(\tau)-2$, $c_{\mathrm{i},22}(\tau)=c(\tau)-200s(\tau)-4c(\tau)s(\tau)$,
$c_{\mathrm{i},23}(\tau)=-s(\tau)-2c(\tau)s(\tau)+299$.

The only exact solution to this example is
\begin{align}
	X^*(\tau)=\begin{bmatrix}
		s(\tau) & c(\tau)& 1\\
		-c(\tau) & -s(\tau)& 0\\
	\end{bmatrix}+\mathrm{i}\begin{bmatrix}
		s(\tau) & c(\tau) & 0\\
		-c(\tau) & -s(\tau) & 1\\
	\end{bmatrix}\in
	\mathbb{C}^{2\times 3}.
\end{align}
\end{example}
\begin{example}\label{eq.example3}
Considering TVSSCME \eqref{eq.sccsme.variant} to verify the evident difference between models Con-CZND1 \eqref{eq.solve.linearerrconcznd1} and Con-CZND2 \eqref{eq.solve.linearerrconcznd2}:
\begin{align} 
	F(\tau)
	=\begin{bmatrix}
		6+s(\tau)	&c(\tau) \\
		c(\tau)	&4+s(\tau)
	\end{bmatrix}
	+\mathrm{i}\begin{bmatrix}
		c(\tau)	& s(\tau) \\
		s(\tau) & c(\tau)
	\end{bmatrix}\in
	\mathbb{C}^{2\times 2},
	\notag
\end{align}
\begin{align}
	A(\tau)
=\begin{bmatrix}
	c(\tau)	&s(\tau) \\
	-s(\tau)	&c(\tau)
\end{bmatrix}
+\mathrm{i}\begin{bmatrix}
	s(\tau)	& c(\tau) \\
	c(\tau) & -s(\tau)
\end{bmatrix}\in
\mathbb{C}^{2\times 2},
	\notag
\end{align}
\begin{align}
	C(\tau)
	=\begin{bmatrix}
			c_{\mathrm{r},11}(\tau) & c_{\mathrm{r},12}(\tau) \\
		c_{\mathrm{r},21}(\tau) & c_{\mathrm{r},22}(\tau) \\
	\end{bmatrix}
	+\mathrm{i}\begin{bmatrix}
			c_{\mathrm{i},11}(\tau) & 	c_{\mathrm{i},12}(\tau) \\
			c_{\mathrm{i},21}(\tau) & 	c_{\mathrm{i},22}(\tau) \\
	\end{bmatrix}\in
	\mathbb{C}^{2\times 2},
	\notag	
\end{align}
where
$c_{\mathrm{r},11}(\tau)=2c^{2}(\tau)-2c(\tau)s(\tau)+6s(\tau)$,
$c_{\mathrm{r},12}(\tau)=4c(\tau)+2c(\tau)s(\tau)-2c^{2}(\tau)$, $c_{\mathrm{r},21}(\tau)=-2s(2\tau)-6c(\tau)+2$, 
$c_{\mathrm{r},22}(\tau)=2s(2\tau)-4s(\tau)-2$ 
and 
$c_{\mathrm{i},11}(\tau)=2c^{2}(\tau)+2c(\tau)s(\tau)+6s(\tau)$, $c_{\mathrm{i},12}(\tau)=4c(\tau)+2c(\tau)s(\tau)+2c^{2}(\tau)$, 
$c_{\mathrm{i},21}(\tau)=-2s(2\tau)-6c(\tau)-2$, $c_{\mathrm{i},22}(\tau)=-2s(2\tau)-4s(\tau)-2$.

The only exact solution to this example is
\begin{align}
	X^*(\tau)=\begin{bmatrix}
		s(\tau) & c(\tau)\\
		-c(\tau) & -s(\tau)\\
	\end{bmatrix}+\mathrm{i}\begin{bmatrix}
		s(\tau) & c(\tau) \\
		-c(\tau) & -s(\tau) \\
	\end{bmatrix}\in
	\mathbb{C}^{2\times 2}.
\end{align}
\end{example}
\begin{remark}
In all three examples, both models take initial random values in the interval $\left [-5,5  \right ] $, and the experimental run time $\tau$ is in the interval $\left [0,10  \right ] $. Sections 4.1 and 4.2 illustrate the validity and convergence of Con-CZND1 \eqref{eq.solve.linearerrconcznd1} model and Con-CZND2 \eqref{eq.solve.linearerrconcznd2} model for matrices $F(\tau)$, $A(\tau)$, and $C(\tau)$ with different matrix dimensions, where $\gamma$ equals 1. Section 4.3 uses residuals and designing special experiments to supplement the continuing models Con-CZND1 \eqref{eq.solve.linearerrconcznd1} and Con-CZND2 \eqref{eq.solve.linearerrconcznd2} error, stability, and some findings, where $\gamma$ equals 10. Since the error matrices of the two models are defined differently, the residuals are uniformly defined $\left \|X(\tau)-X^*(\tau)   \right \|_{\mathrm{F}}$, where $\left \|\cdot   \right \|_{\mathrm{F}}$ stands for Frobenius norm.
\end{remark}
\subsection{Con-CZND1 model}
\begin{figure}[htbp]\centering
	\subfigure[]{\includegraphics[width=0.49\columnwidth]{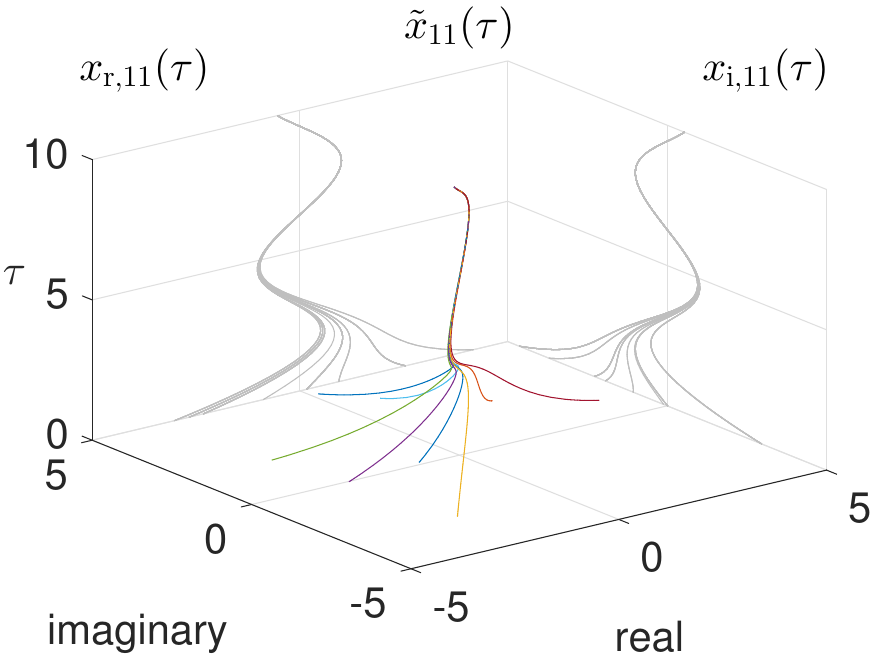}}
	\subfigure[]{\includegraphics[width=0.49\columnwidth]{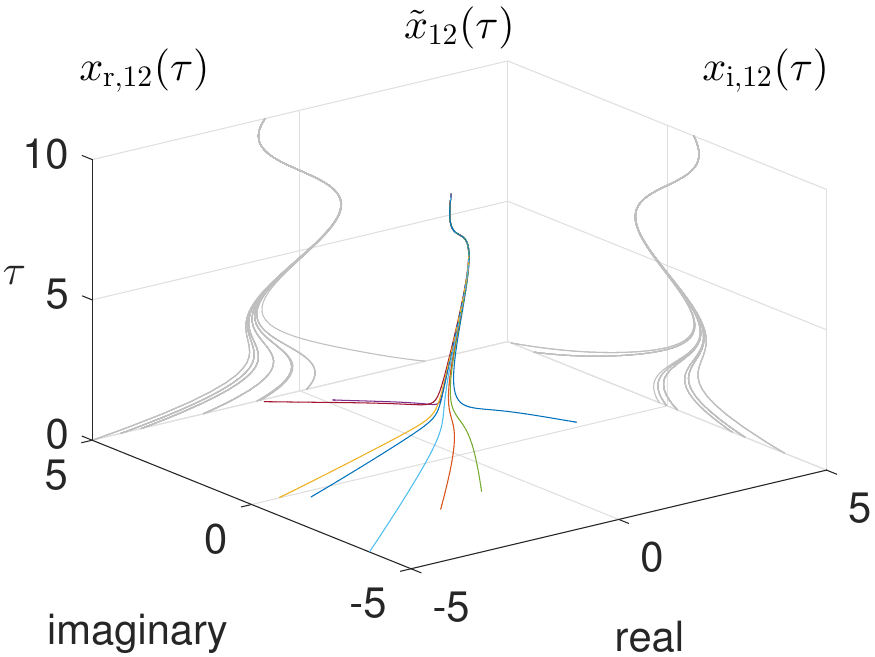}}
	\subfigure[]{\includegraphics[width=0.49\columnwidth]{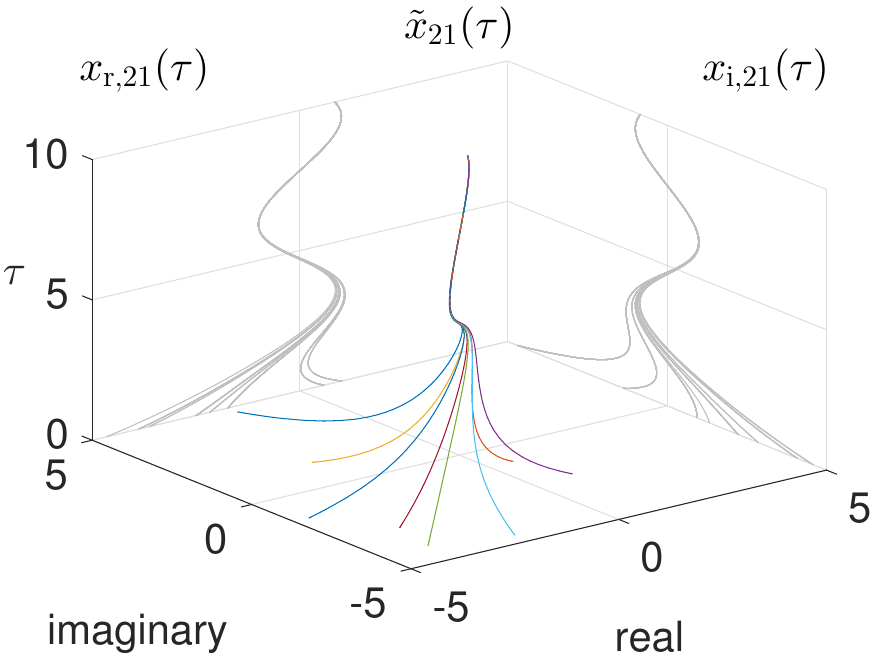}}
	\subfigure[]{\includegraphics[width=0.49\columnwidth]{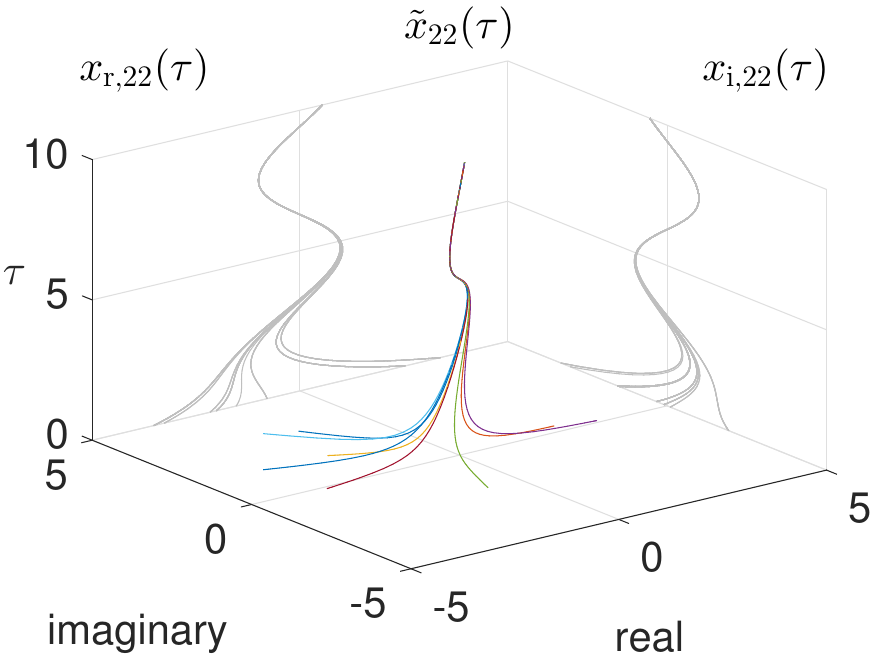}}
	\subfigure[]{\includegraphics[width=0.49\columnwidth]{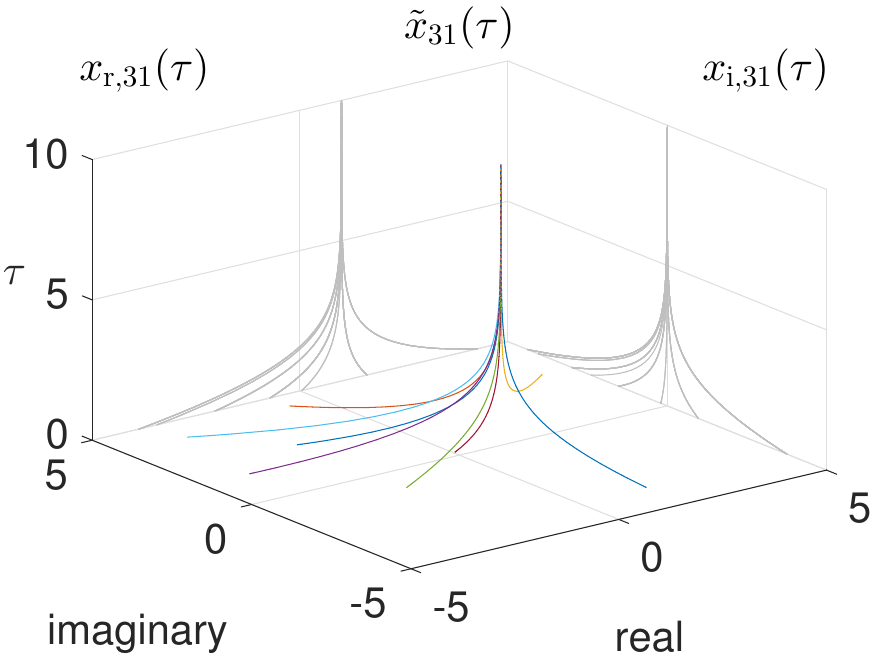}}
	\subfigure[]{\includegraphics[width=0.49\columnwidth]{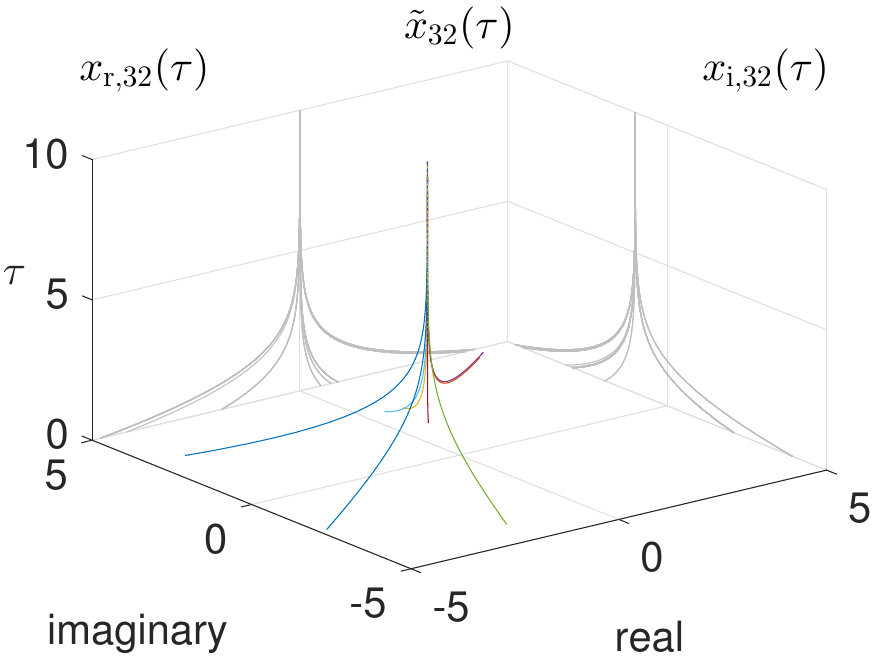}}
	\caption{Solution $X(\tau)$ computed by Con-CZND1 \eqref{eq.solve.linearerrconcznd1} model in Example \ref{eq.example1}.}
	\label{fig.e1.Con-CZND1.solve}
\end{figure}

\begin{figure}[htbp]\centering
	\subfigure[]{\includegraphics[width=0.49\columnwidth]{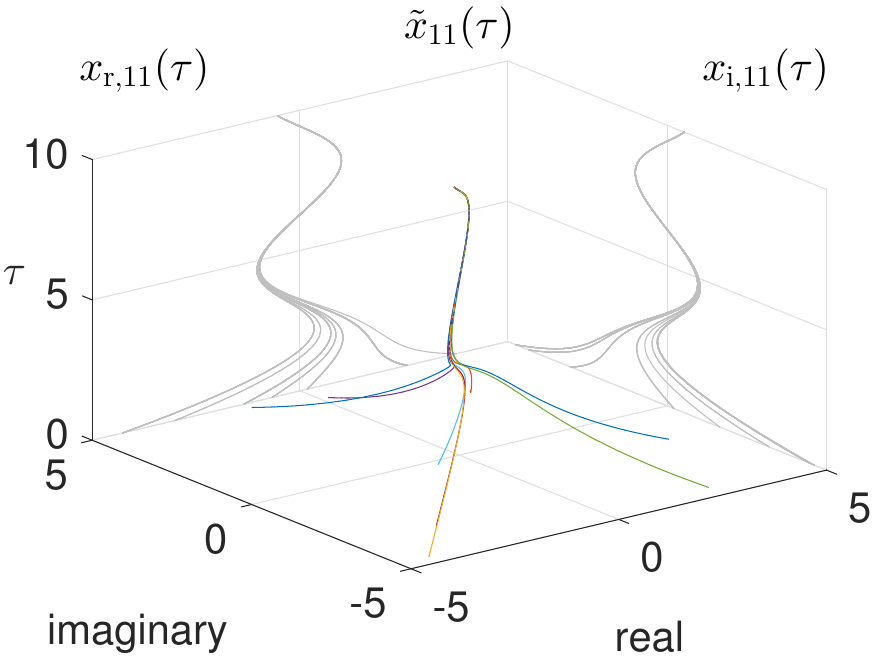}}
	\subfigure[]{\includegraphics[width=0.49\columnwidth]{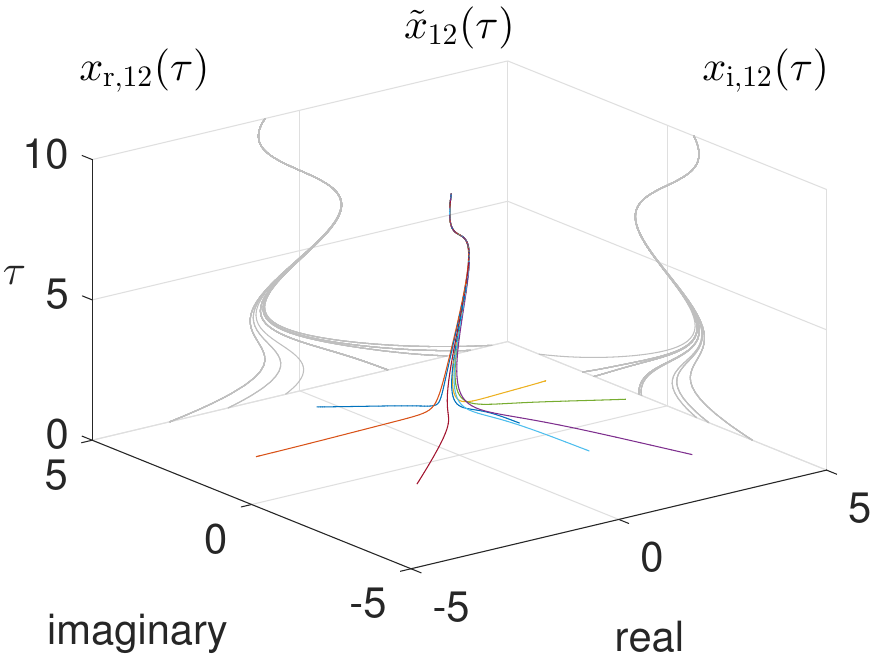}}
	\subfigure[]{\includegraphics[width=0.49\columnwidth]{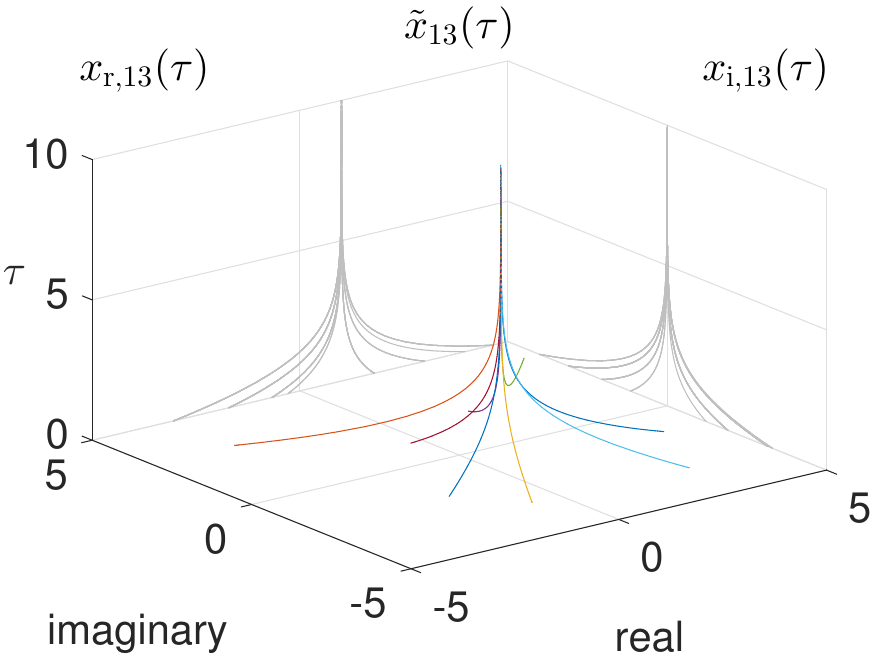}}
	\subfigure[]{\includegraphics[width=0.49\columnwidth]{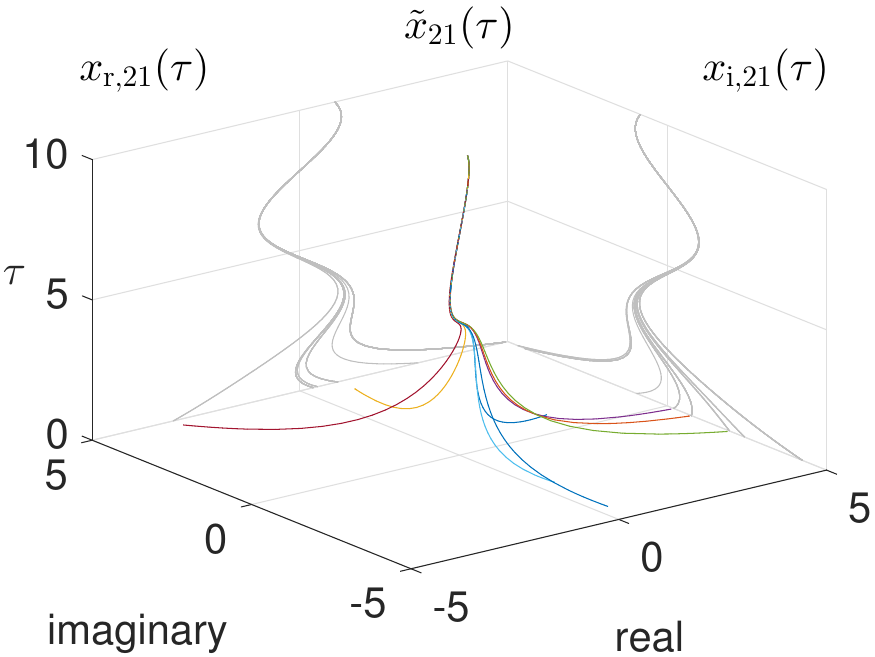}}
	\subfigure[]{\includegraphics[width=0.49\columnwidth]{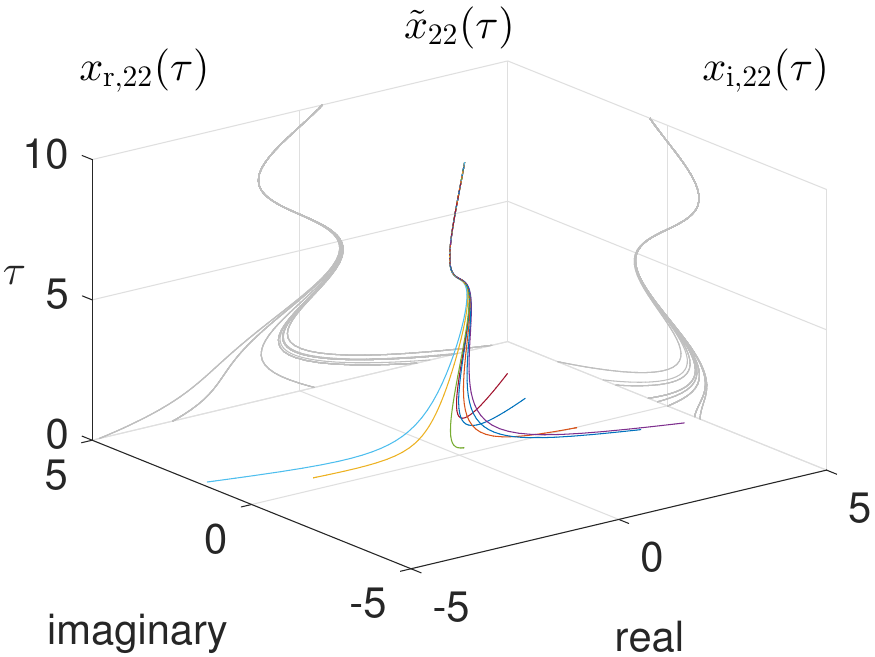}}
	\subfigure[]{\includegraphics[width=0.49\columnwidth]{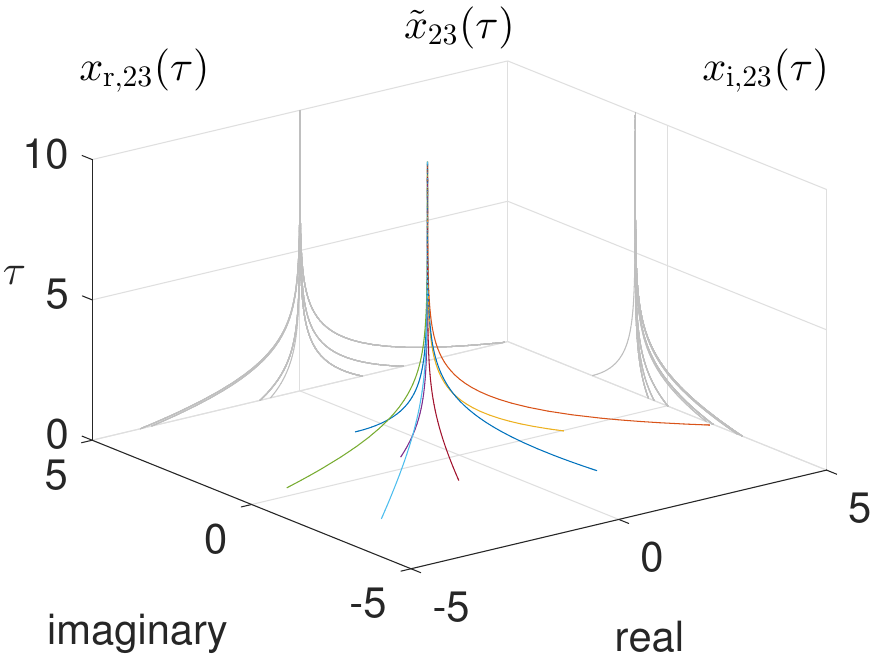}}
	\caption{Solution $X(\tau)$ computed by Con-CZND1 \eqref{eq.solve.linearerrconcznd1} model in Example \ref{eq.example2}.}
	\label{fig.e2.Con-CZND1.solve}
\end{figure}

Using the ode45
\cite{zhangMoreNewtonIterations2010} 
function in MATLAB, Con-CZND1 \eqref{eq.solve.linearerrconcznd1} model is executed, where $\gamma$ equals 1, the solution results of Example \ref{eq.example1} and Example \ref{eq.example2} are shown in 
Figs. \ref{fig.e1.Con-CZND1.solve} and \ref{fig.e2.Con-CZND1.solve}, 
where the grey lines projected on the corresponding planes represent the solving elements corresponding to the real and imaginary matrices.
\begin{figure}[htbp]\centering
	\subfigure[]{\includegraphics[width=0.7\columnwidth]{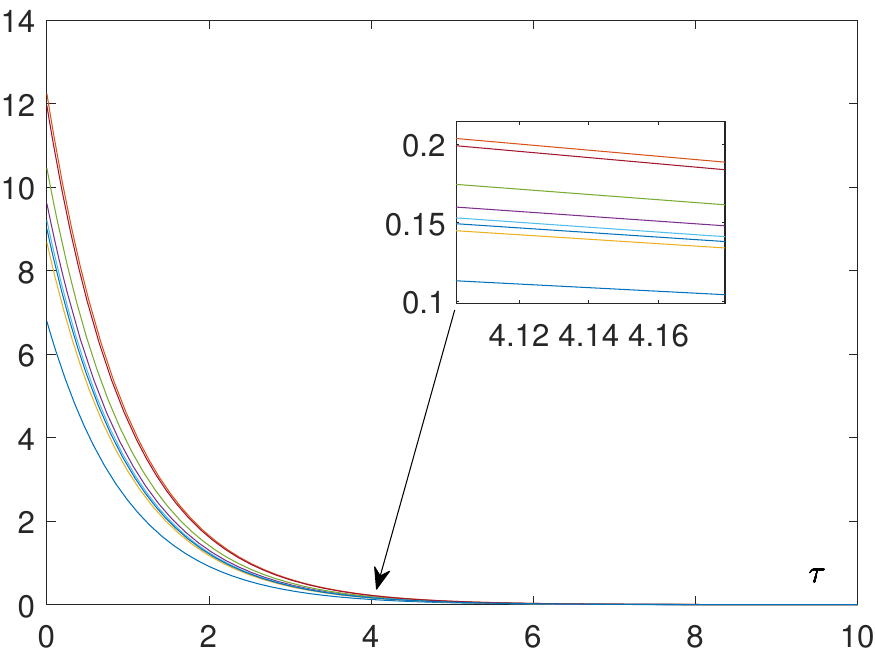}\label{fig.e1.Con-CZND1.normerror.normal}}
	\subfigure[]{\includegraphics[width=0.7\columnwidth]{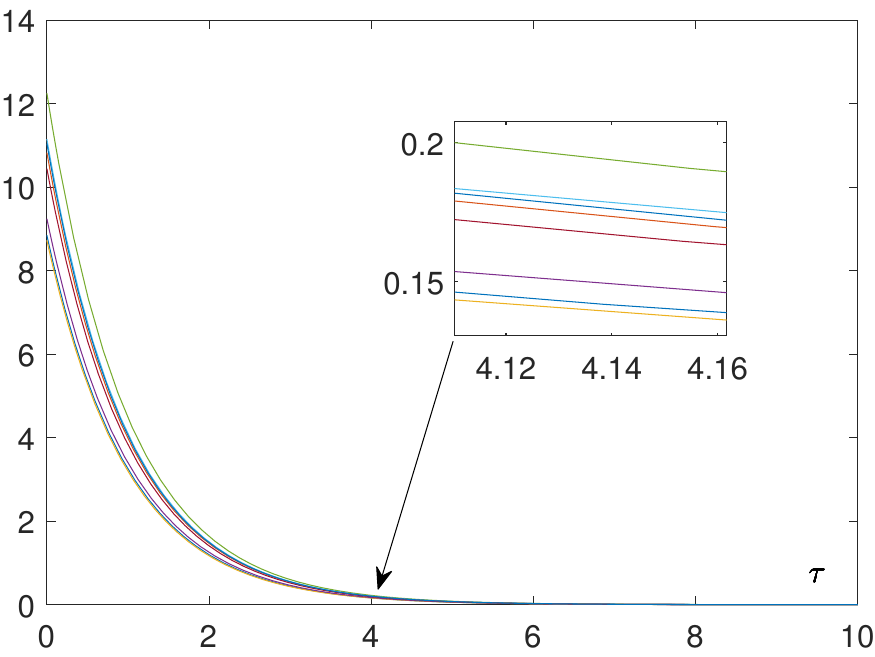}\label{fig.e2.Con-CZND1.normerror.normal}}
\caption{$\left \|X(\tau)-X^*(\tau)   \right \|_{\mathrm{F}}$ computed by Con-CZND1 \eqref{eq.solve.linearerrconcznd1} model in Examples \ref{eq.example1} and \ref{eq.example2}. \subref{fig.e1.Con-CZND1.normerror.normal} Example \ref{eq.example1}. \subref{fig.e2.Con-CZND1.normerror.normal} Example \ref{eq.example2}.}
\end{figure}

It can be seen that the model solution $X(\tau)$ uniformly and essentially matches the target solution $X^*(\tau)$ as time goes by, while the time-variant and time-invariant elements in the solutions of the two examples are one-to-one with the corresponding time-variant and time-invariant elements in the target solution, and as can be seen from Figs. \ref{fig.e1.Con-CZND1.normerror.normal} and \ref{fig.e2.Con-CZND1.normerror.normal}, the residuals of Con-CZND1 \eqref{eq.solve.linearerrconcznd1} model in Examples \ref{eq.example1} and \ref{eq.example2} quickly converge toward zero, which verifies the validity of Con-CZND1 \eqref{eq.solve.linearerrconcznd1} model.
\subsection{Con-CZND2 model} 
\begin{figure}[htbp]\centering
	\subfigure[]{\includegraphics[width=0.49\columnwidth]{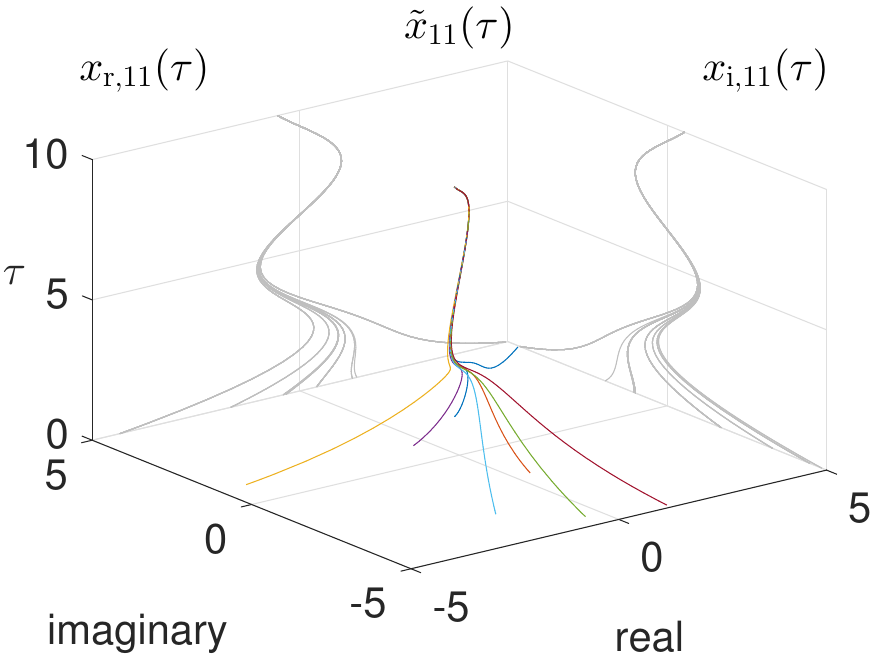}}
	\subfigure[]{\includegraphics[width=0.49\columnwidth]{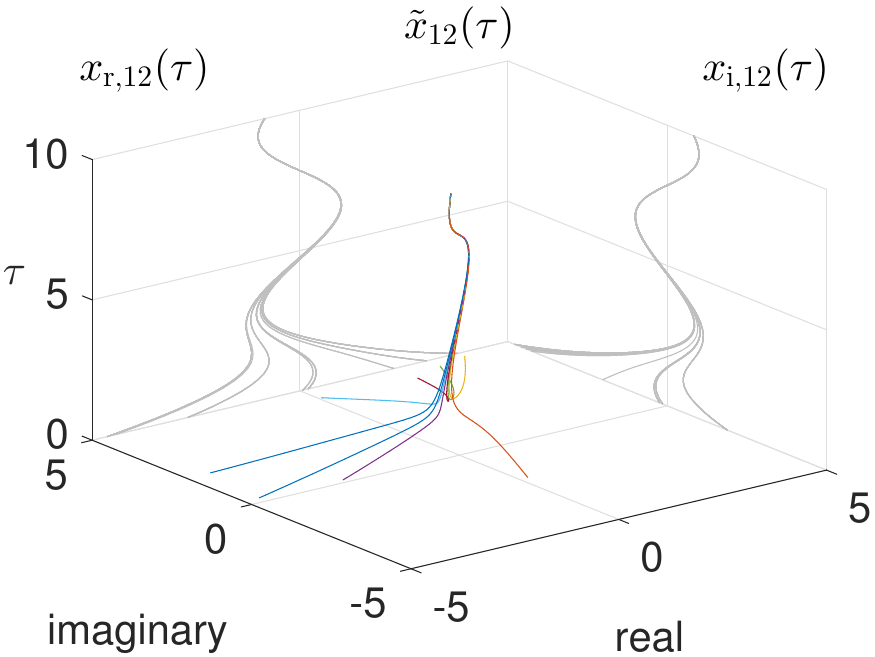}}
	\subfigure[]{\includegraphics[width=0.49\columnwidth]{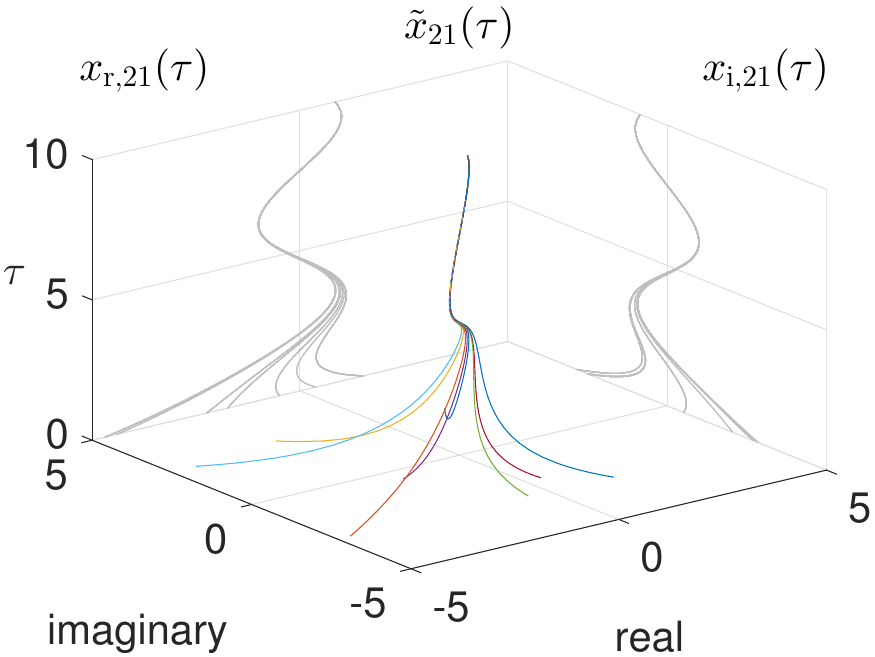}}
	\subfigure[]{\includegraphics[width=0.49\columnwidth]{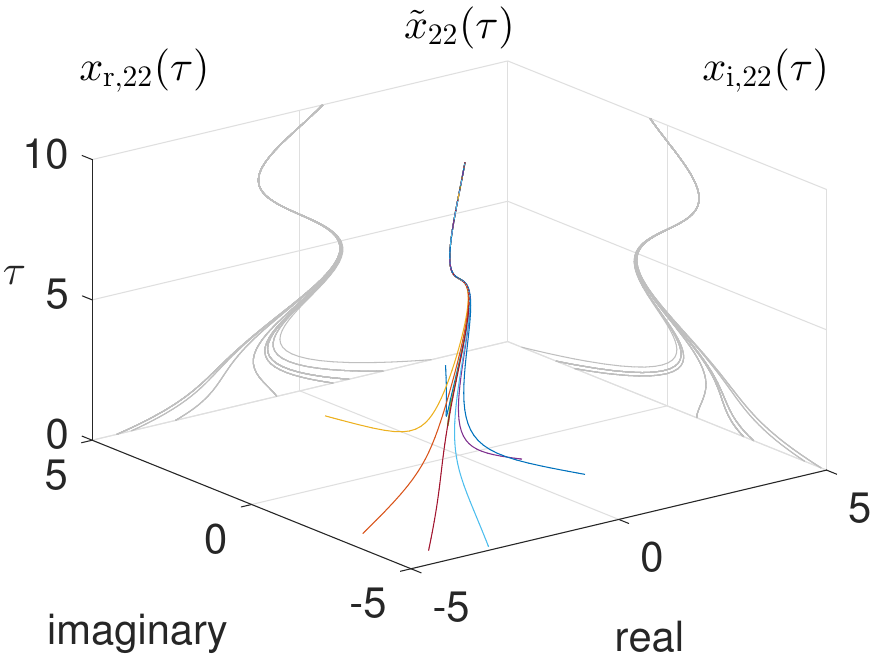}}
	\subfigure[]{\includegraphics[width=0.49\columnwidth]{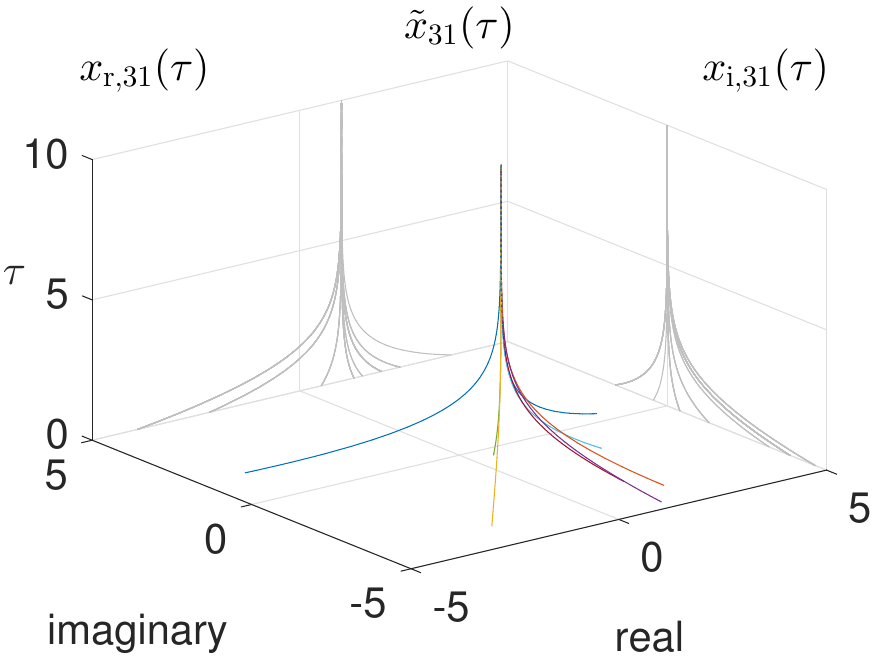}}
	\subfigure[]{\includegraphics[width=0.49\columnwidth]{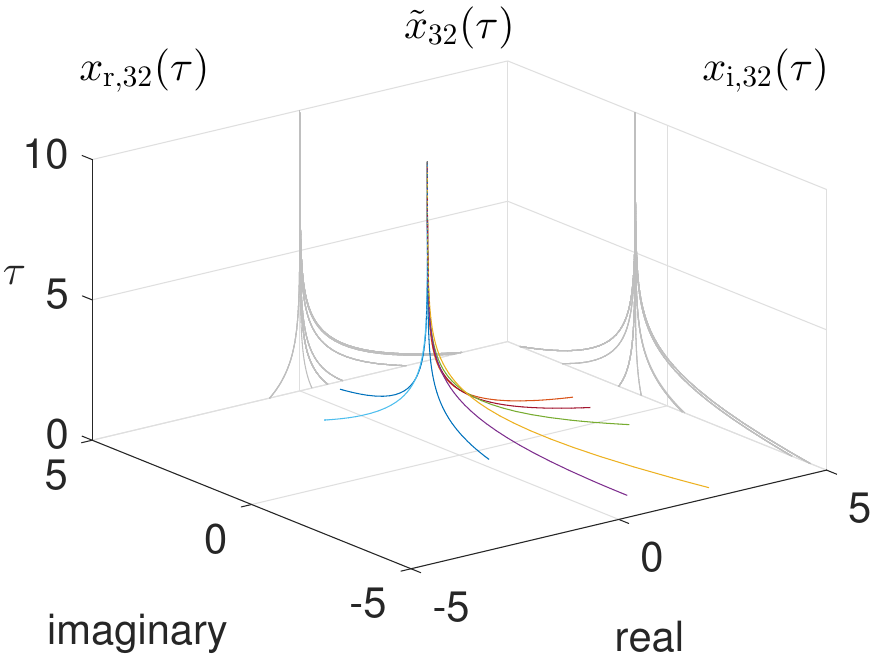}}
	\caption{Solution $X(\tau)$ computed by Con-CZND2 \eqref{eq.solve.linearerrconcznd2} model in Example \ref{eq.example1}.}
	\label{fig.e1.Con-CZND2.solve}
\end{figure}

\begin{figure}[htbp]\centering
	\subfigure[]{\includegraphics[width=0.49\columnwidth]{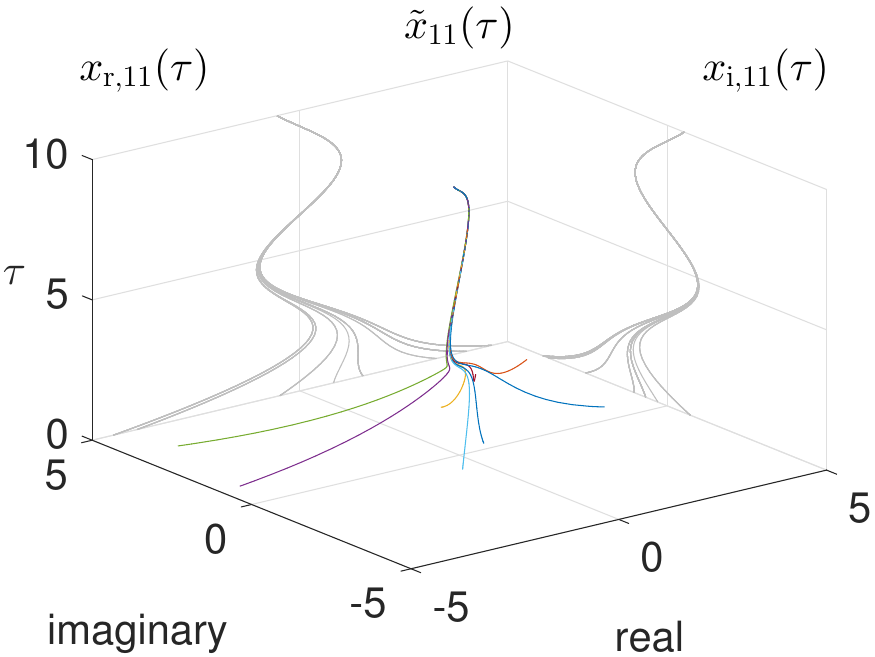}}
	\subfigure[]{\includegraphics[width=0.49\columnwidth]{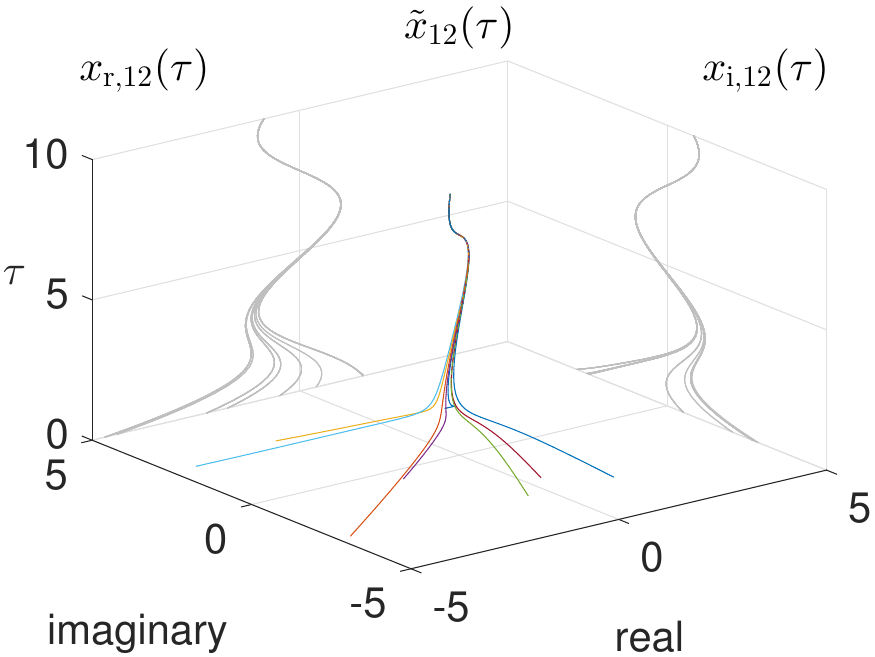}}
	\subfigure[]{\includegraphics[width=0.49\columnwidth]{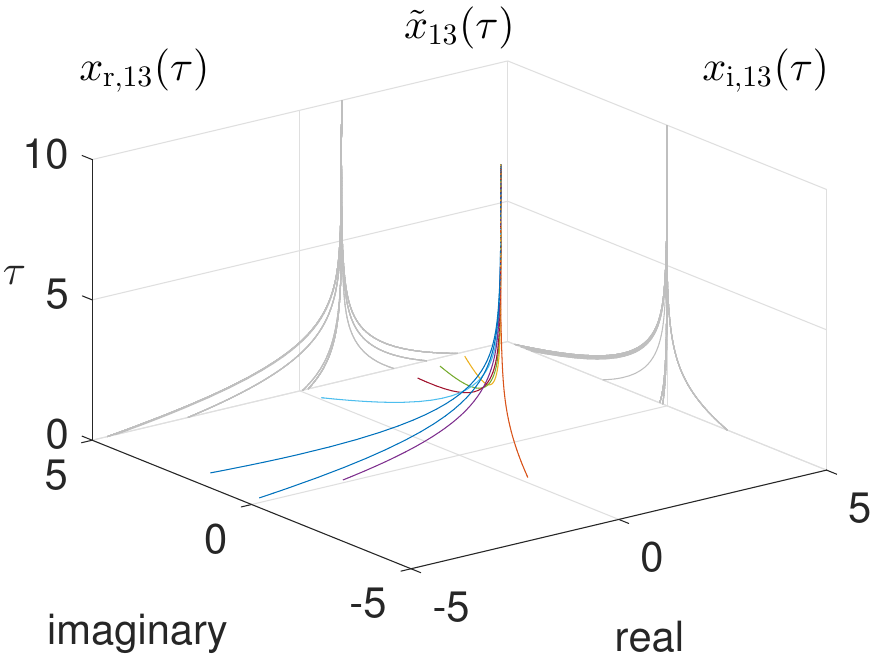}}
	\subfigure[]{\includegraphics[width=0.49\columnwidth]{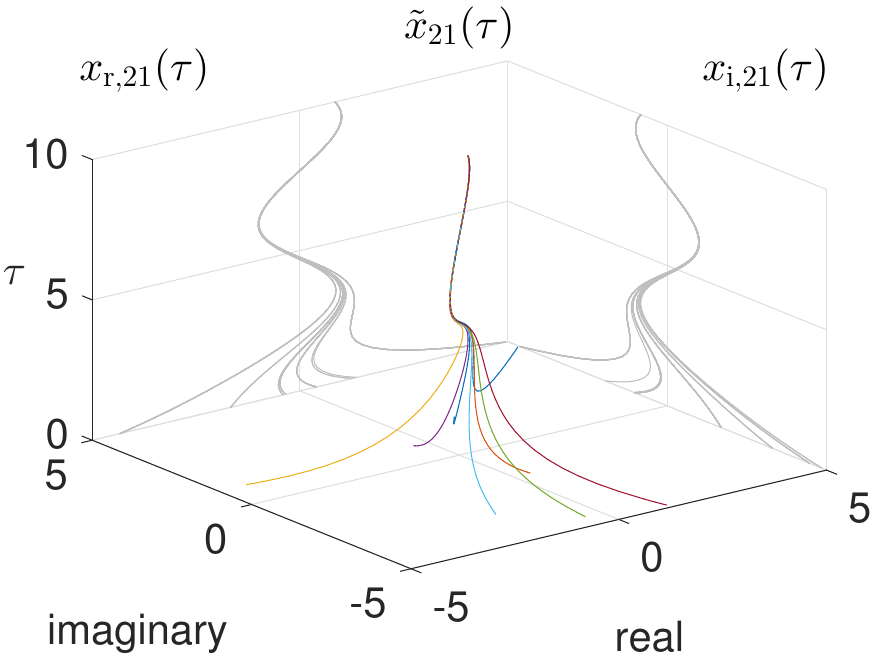}}
	\subfigure[]{\includegraphics[width=0.49\columnwidth]{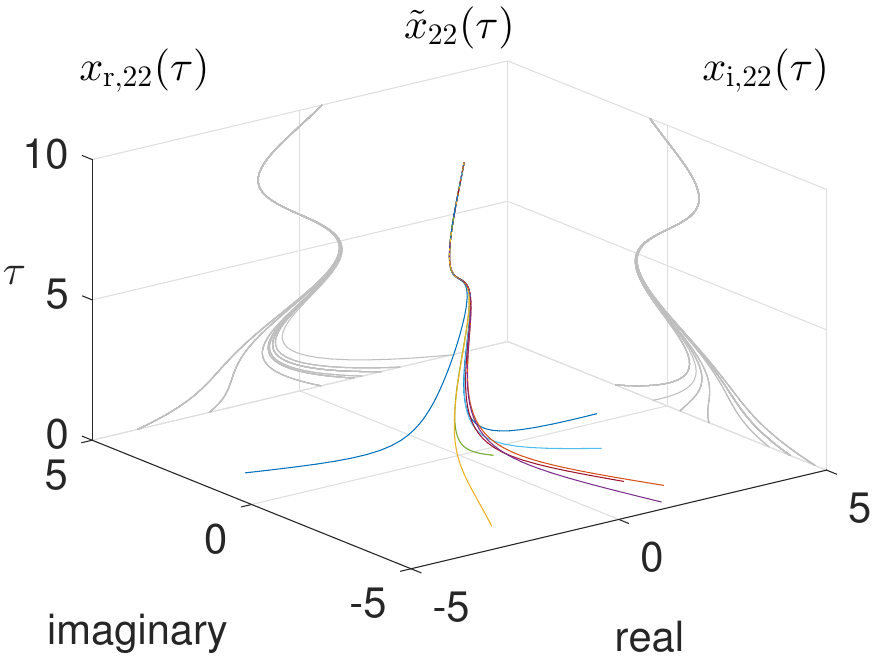}}
	\subfigure[]{\includegraphics[width=0.49\columnwidth]{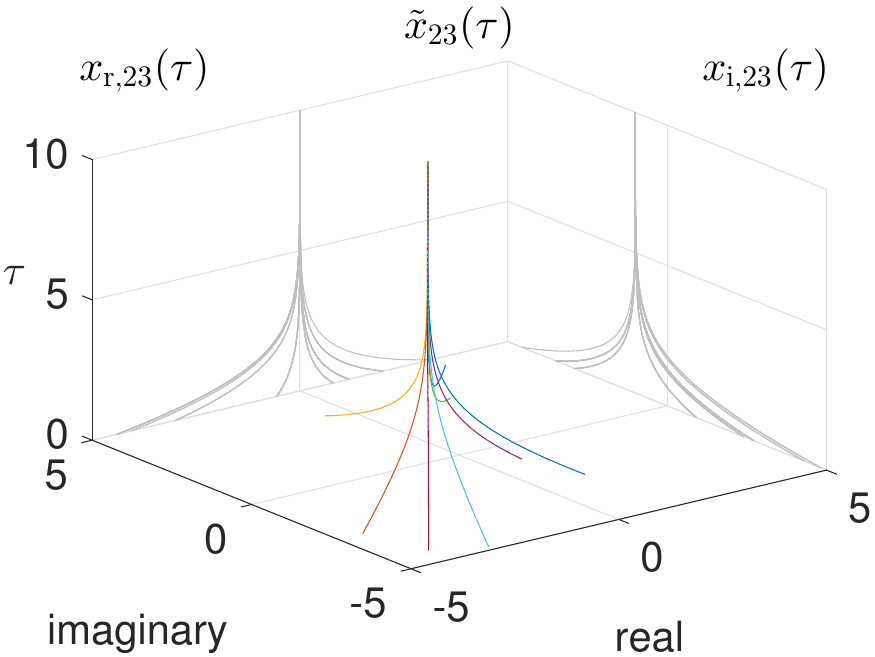}}
	\caption{Solution $X(\tau)$ computed by Con-CZND2 \eqref{eq.solve.linearerrconcznd2} model in Example \ref{eq.example2}.}
	\label{fig.e2.Con-CZND2.solve}
\end{figure}

As in the previous subsection, Con-CZND2 \eqref{eq.solve.linearerrconcznd2} model is executed using the ode45
\cite{zhangMoreNewtonIterations2010}
function in MATLAB, and the results of solving Example \ref{eq.example1} and Example \ref{eq.example2} where $\gamma$ equals 1 are shown in Figs. \ref{fig.e1.Con-CZND2.solve} and \ref{fig.e2.Con-CZND2.solve}, where the grey lines projected on the plane represent the same meaning as in the previous subsection.

\begin{figure}[htbp]\centering
	\subfigure[]{\includegraphics[width=0.7\columnwidth]{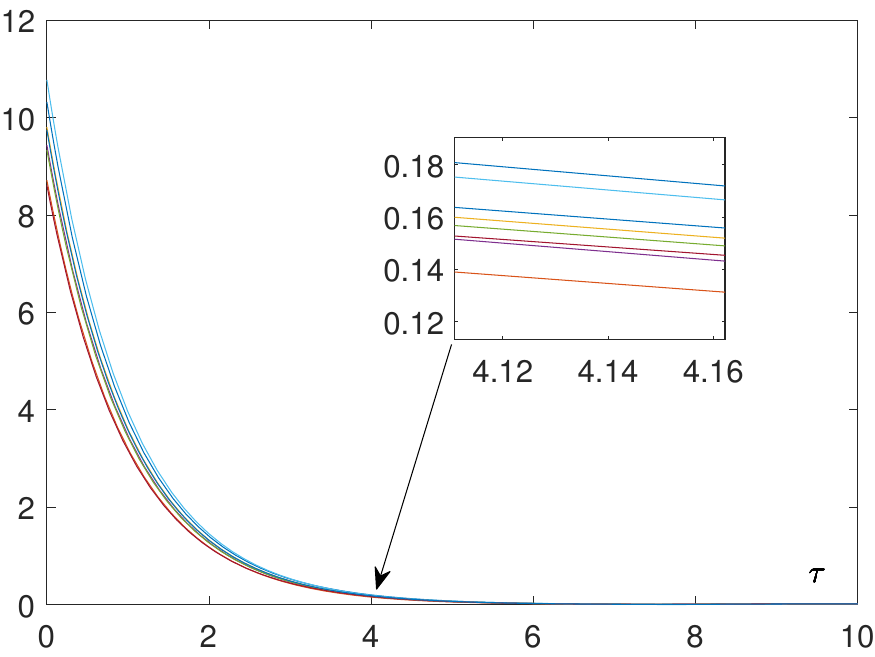}\label{fig.e1.Con-CZND2.normerror.normal}}
	\subfigure[]{\includegraphics[width=0.7\columnwidth]{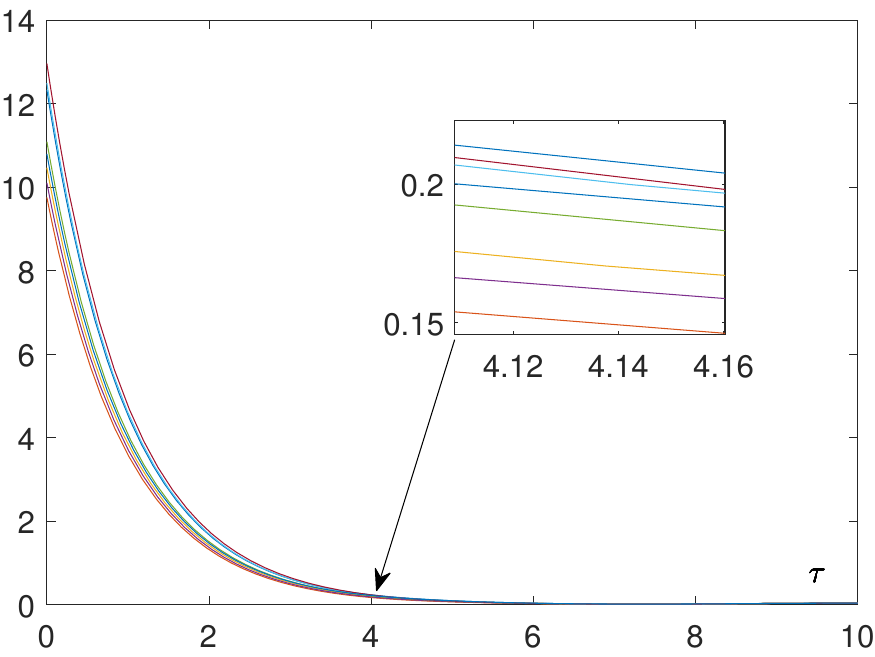}\label{fig.e2.Con-CZND2.normerror.normal}}
	\caption{$\left \|X(\tau)-X^*(\tau)   \right \|_{\mathrm{F}}$ computed by Con-CZND2 \eqref{eq.solve.linearerrconcznd2} model in Examples \ref{eq.example1} and \ref{eq.example2}. \subref{fig.e1.Con-CZND2.normerror.normal} Example \ref{eq.example1}. \subref{fig.e2.Con-CZND2.normerror.normal} Example \ref{eq.example2}.}
\end{figure}

It can be seen that the model solutions $X(\tau)$ uniformly and essentially match the target solution $X^*(\tau)$ with time, while the time-variant and time-invariant elements in the solutions of the two examples are one-to-one with the corresponding time-variant and time-invariant elements in the target solution, and it can be seen from Figs. \ref{fig.e1.Con-CZND2.normerror.normal} and \ref{fig.e2.Con-CZND2.normerror.normal} that the residuals of Con-CZND2 \eqref{eq.solve.linearerrconcznd2} model in Examples \ref{eq.example1} and \ref{eq.example2} quickly converge toward zero.

However, above two examples do not clearly show the difference for both models, so the logarithmic residual $\left \|X(\tau)-X^*(\tau)   \right \|_{\mathrm{F}}$ trajectories and Example \ref{eq.example3} are added.
\subsection{Con-CZND1 model vs Con-CZND2 model by comparing error and stability}
\begin{figure}[htbp]\centering
	\includegraphics[width=0.7\columnwidth]{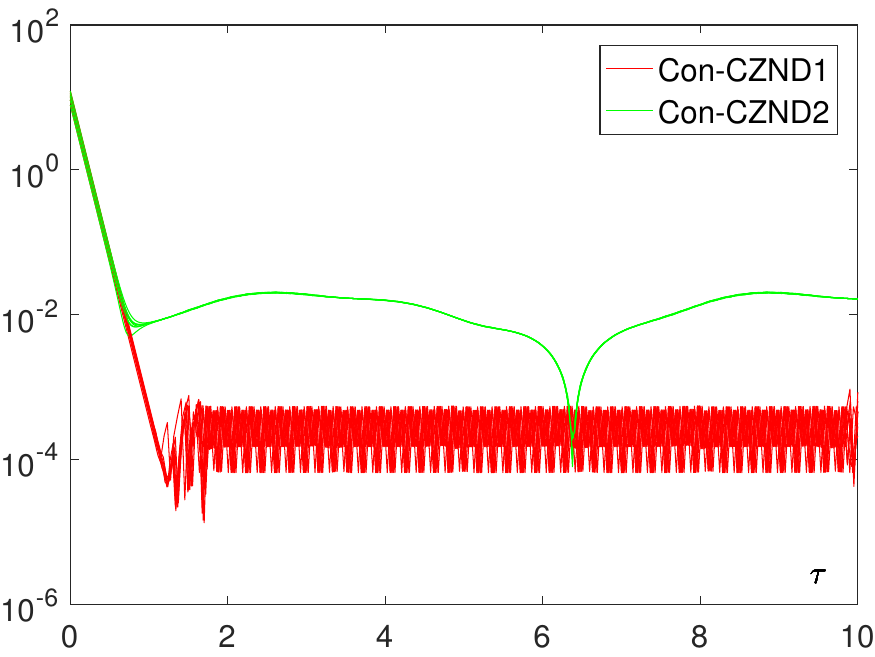}
	\caption{Logarithmic residual $\left \|X(\tau)-X^*(\tau)   \right \|_{\mathrm{F}}$ trajectories computed by Con-CZND1 \eqref{eq.solve.linearerrconcznd1} model vs. Con-CZND2 \eqref{eq.solve.linearerrconcznd2} model in Example 1.
		}
	\label{fig.e1.Con-CZND1 vs Con-CZND2.normerror.logarithmic}
\end{figure}

\begin{figure}[htbp]\centering
	\includegraphics[width=0.7\columnwidth]{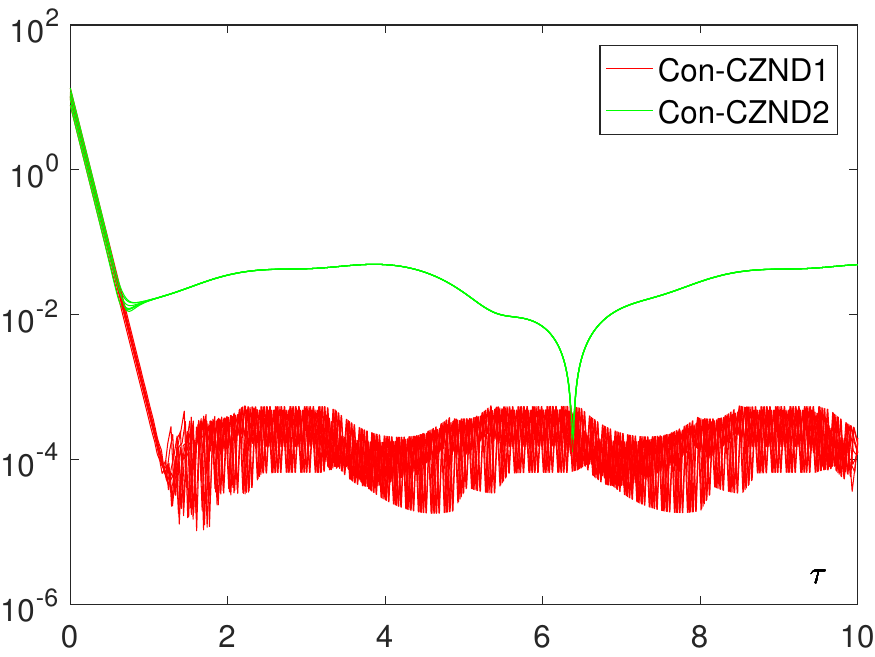}
	\caption{Logarithmic residual $\left \|X(\tau)-X^*(\tau)   \right \|_{\mathrm{F}}$ trajectories computed by Con-CZND1 \eqref{eq.solve.linearerrconcznd1} model vs. Con-CZND2 \eqref{eq.solve.linearerrconcznd2} model in Example 2.}
	\label{fig.e2.Con-CZND1 vs Con-CZND2.normerror.logarithmic}
\end{figure}

Where $\gamma$ equals 10, the logarithmic residual $\left \|X(\tau)-X^*(\tau)   \right \|_{\mathrm{F}}$ trajectories of Con-CZND1
\eqref{eq.solve.linearerrconcznd1} model and Con-CZND2 \eqref{eq.solve.linearerrconcznd2} model in Examples \ref{eq.example1} and \ref{eq.example2} are shown in Figs. \ref{fig.e1.Con-CZND1 vs Con-CZND2.normerror.logarithmic} and \ref{fig.e2.Con-CZND1 vs Con-CZND2.normerror.logarithmic}.

In Figs. \ref{fig.e1.Con-CZND1 vs Con-CZND2.normerror.logarithmic} and \ref{fig.e2.Con-CZND1 vs Con-CZND2.normerror.logarithmic}, Con-CZND1 \eqref{eq.solve.linearerrconcznd1} model has better accuracy than Con-CZND2 \eqref{eq.solve.linearerrconcznd2} model, with the residuals stabilized in a range. It is found that Con-CZND2 \eqref{eq.solve.linearerrconcznd2} model is strongly influenced by the absolute values of the main diagonal elements of the matrix $W_{\mathrm{M2}}(\tau)\in
\mathbb{R}^{2mn\times 2mn}$. In order to verify that Con-CZND2 \eqref{eq.solve.linearerrconcznd2} model is affected by the main diagonal predominance of $W_{\mathrm{M2}}(\tau)$, Example \ref{eq.example3} is designed: i.e., weakening the main diagonal dominance of $W_{\mathrm{M2}}(\tau)$ by weakening the main diagonal dominance of the real matrix of the square matrix $F(\tau)\in
\mathbb{C}^{n\times n}$, where the red dotted lines are of the corresponding exact solution $X^*(\tau)\in
\mathbb{C}^{m\times n}$.

\begin{figure}[htbp]\centering
	\subfigure[]{\includegraphics[width=0.7\columnwidth]{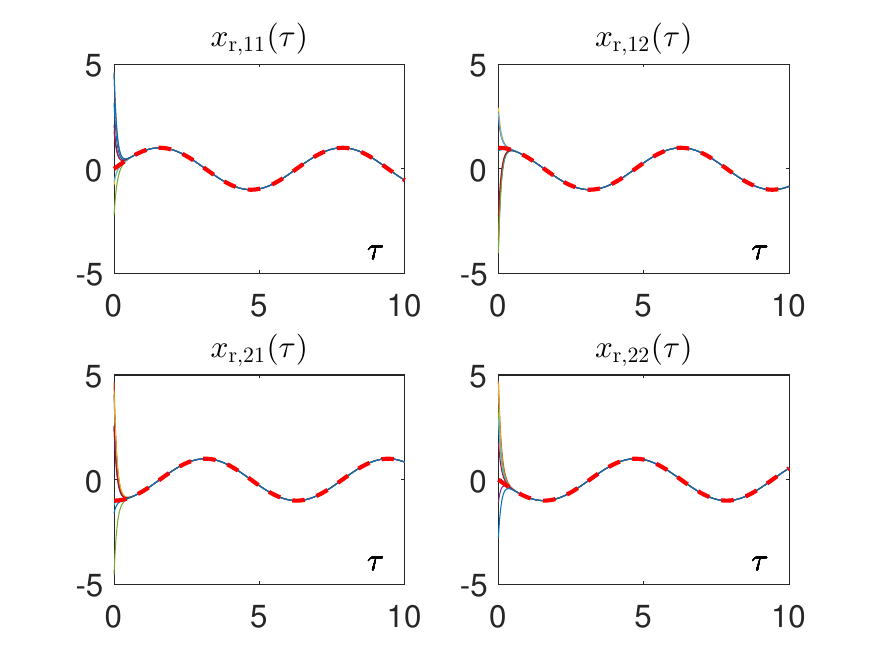}}
	\subfigure[]{\includegraphics[width=0.7\columnwidth]{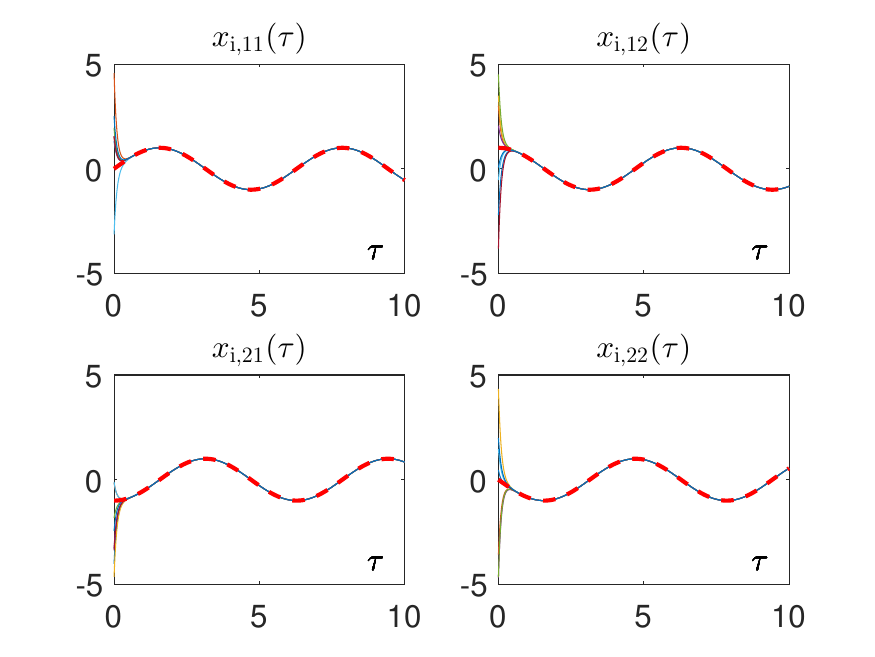}}
	\caption{Solution $X(\tau)$ computed by Con-CZND1 \eqref{eq.solve.linearerrconcznd1} model and exact solution $X^*(\tau)$ in Example \ref{eq.example3}.}
	\label{fig.e3.Con-CZND1.solve}
\end{figure}

\begin{figure}[htbp]\centering
	\subfigure[]{\includegraphics[width=0.7\columnwidth]{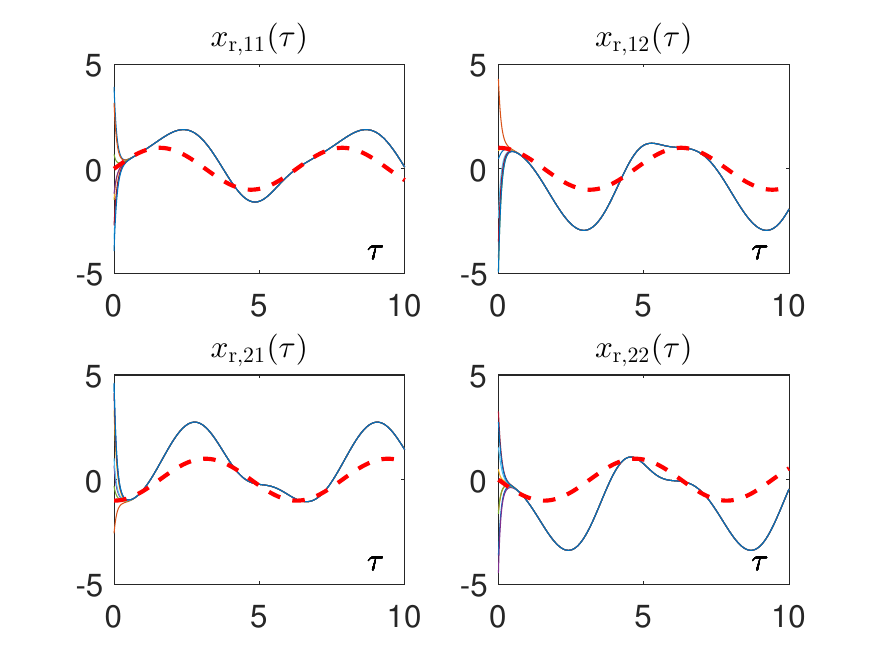}}
	\subfigure[]{\includegraphics[width=0.7\columnwidth]{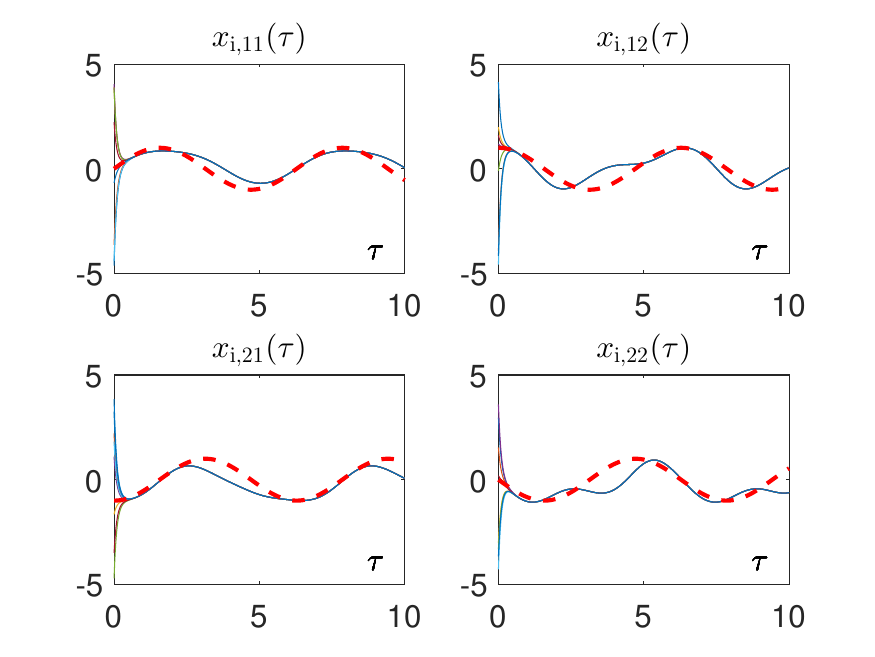}}
	\caption{Solution $X(\tau)$ computed by Con-CZND2 \eqref{eq.solve.linearerrconcznd2} model and exact solution $X^*(\tau)$ in Example \ref{eq.example3}.}
	\label{fig.e3.Con-CZND2.solve}
\end{figure}

\begin{figure}[htbp]\centering
	\subfigure[]{\includegraphics[width=0.7\columnwidth]{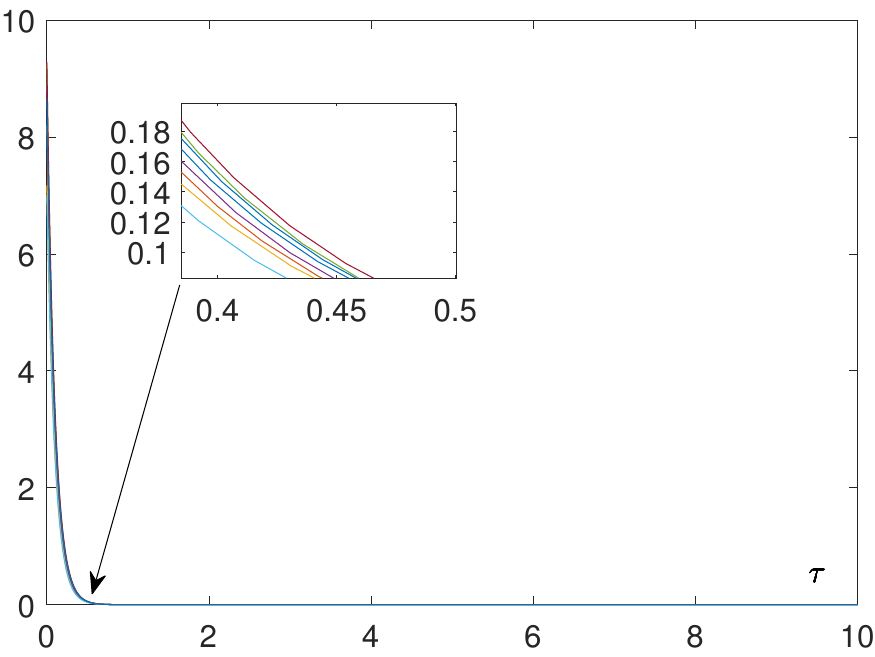}\label{fig.e3.Con-CZND1.normerror.normal}}
	\subfigure[]{\includegraphics[width=0.7\columnwidth]{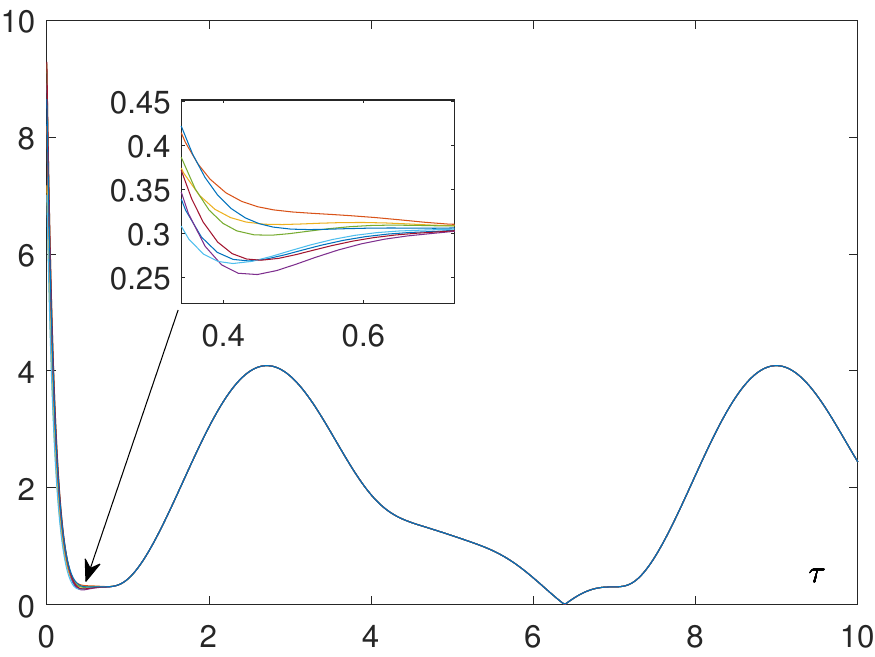}\label{fig.e3.Con-CZND2.normerror.normal}}
	\caption{$\left \|X(\tau)-X^*(\tau)   \right \|_{\mathrm{F}}$ computed by Con-CZND1 \eqref{eq.solve.linearerrconcznd1} model vs. Con-CZND2 \eqref{eq.solve.linearerrconcznd2} model in Example 3.
	\subref{fig.e3.Con-CZND1.normerror.normal} Con-CZND1 \eqref{eq.solve.linearerrconcznd1} model. \subref{fig.e3.Con-CZND2.normerror.normal} Con-CZND2 \eqref{eq.solve.linearerrconcznd2} model.}
	\label{fig.e3.Con-CZND1 vs Con-CZND2.normerror.normal}
\end{figure}

\begin{figure}[htbp]\centering
	\includegraphics[width=0.7\columnwidth]{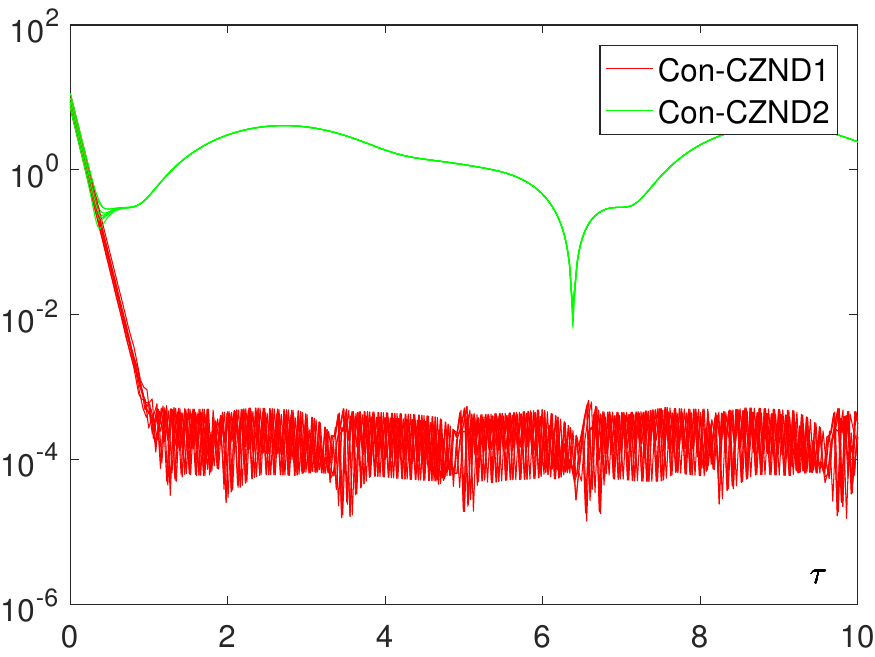}
	\caption{Logarithmic residual $\left \|X(\tau)-X^*(\tau)   \right \|_{\mathrm{F}}$ trajectories computed by Con-CZND1 \eqref{eq.solve.linearerrconcznd1} model vs. Con-CZND2 \eqref{eq.solve.linearerrconcznd2} model in Example 3.}
	\label{fig.e3.Con-CZND1 vs Con-CZND2.normerror.logarithmic}
\end{figure}

In Figs. \ref{fig.e3.Con-CZND1.solve}
through
\ref{fig.e3.Con-CZND1 vs Con-CZND2.normerror.logarithmic},
it can be seen that the residuals of Con-CZND1 \eqref{eq.solve.linearerrconcznd1} model are not affected by the dominance of the main diagonal of the real matrix of the square matrix $F(\tau)\in
\mathbb{C}^{n\times n}$. And the logarithmic residual $\left \|X(\tau)-X^*(\tau)   \right \|_{\mathrm{F}}$ trajectory computed by Con-CZND1 \eqref{eq.solve.linearerrconcznd1} model in Example \ref{eq.example3} is still stabilized in the same intervals as in Example \ref{eq.example1} and Example \ref{eq.example2}, since the error matrix $E_{\mathrm{M1}}(\tau)\in
\mathbb{C}^{m\times n}$, has a total of $m\times n$ elements. But Con-CZND2 \eqref{eq.solve.linearerrconcznd2} model shows a large shock in the computation, since the error matrix $E_{\mathrm{M2}}(\tau)\in\mathbb{R}^{2mn\times 1}$, defined in terms of the matrix transformation, has a total of $2\times m\times n$ elements, which is equivalent to the dimensionality becoming larger according to the knowledge of the matrix computations 
\cite{golubMatrixComputations4th2013}.
Example \ref{eq.example3} validates the previous statement and highlights that the complex field ZND should be used directly for this type of matrix equation.
\section{Conclusion}
In this paper, based on an investigation of the earliest time-variant CCME, TVSSCME \eqref{eq.sccsme.variant} is studied. Firstly, the knowledge about the vectorization and Kronecker product under the relevant complex field is supplemented. Secondly, Con-CZND1 \eqref{eq.solve.linearerrconcznd1} model, which deals with the error term directly in the complex field, and Con-CZND2 \eqref{eq.solve.linearerrconcznd2} model, which deals with the error matrix in the mapping real field, are proposed. Finally, through theoretical analyses and experimental results, the advantages of handling TVSSCME \eqref{eq.sccsme.variant} in the complex field are highlighted, and the theory of ZND in the complex field is further refined. Subsequently, we will continue to conduct further research based on this study to improve and supplement other relevant knowledge on CCME.


\section*{Acknowledgment}
This work is aided by the Project Supported by the Guangzhou Science and Technology Program (with number 2023E04J1240).

The authors would like to sincerely thank the editors and anonymous
reviewers for their time and efforts spent in handling the paper, as
well as for providing many constructive comments and suggestions for
improving further the presentation and quality of this paper.

\section*{References}

\bibliographystyle{elsarticle-num}

\bibliography{bib-example}

\newpage

\begin{figure}
	\includegraphics[width=0.13\columnwidth]{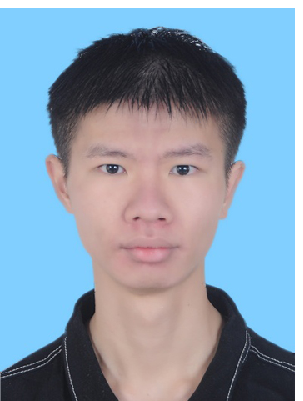}\noindent
	{\bf Jiakuang He} graduated from Guangzhou University of Chinese Medicine, Guangzhou, China, in 2019. He received his dual B.S. degree in Biotechnology as well as Computer Science and Technology from 2019 to 2023, Zhongkai University of Agricultural Engineering, Guangzhou, China. He is now pursuing the M.S. degree in Electronic Information, Zhongkai University of Agricultural Engineering, Guangzhou, China. His current research interests include the intersection of mathematics, chemistry, biology, medicine, and computing.
\end{figure}

\begin{figure}
	\includegraphics[width=0.13\columnwidth]{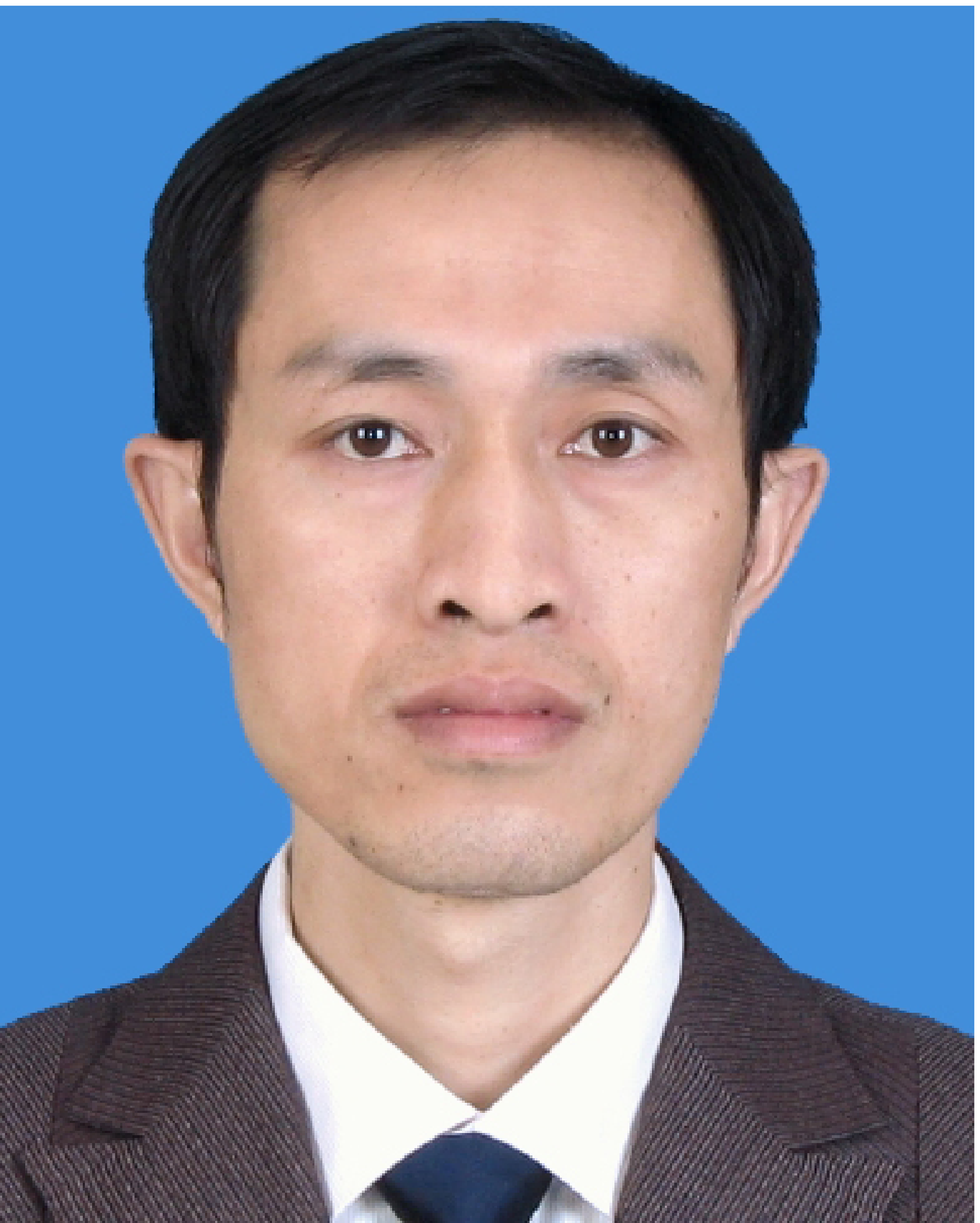}\noindent
	{\bf Dongqing Wu} received the M.S. degree in computer graphics from the Institute of Industrial Design and Graphics, South China University of Technology, Guangzhou, China, in 2005, and the Ph.D. degree in mechanical engineering from the School of Electromechanical Engineering, Guangdong University of Technology, Guangzhou, China, in 2019. He is currently a professor with the School of Mathematics and Data Science, Zhongkai University of Agriculture and Engineering, Guangzhou, China. His current research interests include neural networks, robotics, and numerical analysis.
\end{figure}

\end{document}